\documentclass{article} 
\usepackage{iclr2027_conference,times}

\usepackage{amsmath,amsfonts,bm}

\def\eqref#1{equation~\ref{#1}}

\def\1{\bm{1}}

\DeclareMathAlphabet{\mathsfit}{\encodingdefault}{\sfdefault}{m}{sl}
\SetMathAlphabet{\mathsfit}{bold}{\encodingdefault}{\sfdefault}{bx}{n}

\usepackage{hyperref}
\usepackage{url}
\usepackage{xspace}
\usepackage{algorithm}
\usepackage{algpseudocode}
\usepackage[T1]{fontenc}
\usepackage{listings}
\usepackage{fontawesome5}
\usepackage{placeins}   

\usepackage{booktabs}
\usepackage{makecell}
\usepackage{graphicx}
\usepackage{pifont}
\usepackage{tikz}
\usepackage{pgfplots}
\usetikzlibrary{patterns.meta}
\pgfplotsset{compat=1.18}
\usepackage{subcaption}
\usepackage{wrapfig}
\usepackage{caption}
\usepackage{titletoc}
\usepackage{tabularx}
\usepackage[table]{xcolor}

\newcolumntype{Y}[1]{>{\hsize=#1\hsize\linewidth=\hsize\raggedright\arraybackslash}X}

\definecolor{lightyellow}{RGB}{255, 249, 196}
\definecolor{cardinal}{HTML}{8C1515}   

\newcommand{\highlight}[1]{%
  \begingroup
  \setlength{\fboxsep}{4pt}%
  \colorbox{lightyellow}{#1}%
  \endgroup
}

\definecolor{seedless}{HTML}{EB6834}   
\definecolor{withseed}{HTML}{2A78D6}   
\definecolor{inkmuted}{HTML}{6B6A63}
\definecolor{gridline}{HTML}{E4E3DC}

\usepgfplotslibrary{groupplots}
\usepackage{amsmath}
\usetikzlibrary{shapes.geometric}
\pgfdeclareplotmark{fstar}{%
  \node[star, star points=5, star point ratio=2.3, inner sep=0pt,
        minimum size=2.4*\pgfplotmarksize, draw=black, line width=0.3pt,
        fill=cSHELDON] {};}

\definecolor{cEvoX}{HTML}{4C72B0}
\definecolor{cArbor}{HTML}{55A868}
\definecolor{cCORAL}{HTML}{8172B2}
\definecolor{cSHELDON}{HTML}{C44E52}

\definecolor{evox}{HTML}{1F77B4}
\definecolor{coral}{HTML}{9467BD}
\definecolor{arbor}{HTML}{2CA02C}

\definecolor{ours}{HTML}{C62828}
\definecolor{sheldon}{HTML}{C62828}       
\definecolor{lightsheldon}{HTML}{F2A7A7}  
\definecolor{direct}{HTML}{E8A200}        
\definecolor{gos}{HTML}{17A2B8}           
\definecolor{gold}{HTML}{E0A400}
\definecolor{goldfill}{HTML}{FFF4C2}
\definecolor{band}{HTML}{EEF1F6}
\definecolor{refgrey}{HTML}{7A7F87}

\definecolor{c1}{HTML}{2A78D6}
\definecolor{c2}{HTML}{EB6834}
\definecolor{c3}{HTML}{1BAF7A}
\definecolor{c4}{HTML}{EDA100}
\definecolor{c5}{HTML}{E87BA4}
\definecolor{ink}{HTML}{1A1A19}
\definecolor{muted}{HTML}{6B6A63}
\definecolor{track}{HTML}{D9D8D0}

\newcommand{\sheldonlogo}{\raisebox{-0.22ex}{\includegraphics[height=1.05em]{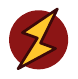}}\hspace{0.08em}}
\newcommand{\logosystem}{\texorpdfstring{\mbox{\sheldonlogo SHELDON}}{SHELDON}\xspace}
\newcommand{\system}{{SHELDON}\xspace}

\title{\texorpdfstring{\sheldonlogo SHELDON}{SHELDON}: Reasoning through Dual Search over Hypothesis and Observation Spaces}

\author{Harshit Joshi\thanks{Equal contribution | \href{https://github.com/stanford-oval/sheldon}{https://github.com/stanford-oval/sheldon}},
Priyank Shethia\footnotemark[1],
\& Monica S. Lam \\
Stanford University, USA\\
\texttt{\{hj, pshethia, lam\}@cs.stanford.edu}}

\newcommand{\Explore}{\textsc{Explore}\xspace}
\newcommand{\Answer}{\textsc{Answer}\xspace}

\newcommand{\LocalTest}{\textsc{CandidateAnalysis}\xspace}
\newcommand{\GlobalTest}{\textsc{ComparativeAnalysis}\xspace}

\newcommand{\revise}{\textsc{Exploit}\xspace}
\newcommand{\Revise}{\textsc{Exploit}\xspace}
\newcommand{\Provisional}{\textsc{Provisional}\xspace}

\newcommand{\wh}{\textit{wH}}
\newcommand{\wo}{\textit{wO}}
\newcommand{\UpdateObservations}{\textsc{UpdateO}}
\newcommand{\status}{\textsc{status}\xspace}
\newcommand{\retained}{\textsc{retained}\xspace}
\newcommand{\setaside}{\textsc{setaside}\xspace}

\definecolor{heatcell}{HTML}{2E8B57}
\newcommand{\heatcell}[2]{\cellcolor{heatcell!#1}#2}

\lstdefinestyle{prompt}{
  basicstyle=\ttfamily\scriptsize, columns=fullflexible, keepspaces=true,
  breaklines=true, breakindent=0pt, postbreak={},
  frame=single, rulecolor=\color{black!30}, backgroundcolor=\color{black!3},
  xleftmargin=2pt, xrightmargin=2pt, aboveskip=4pt, belowskip=8pt,
  escapeinside={(*@}{@*)}, upquote=true, showstringspaces=false,
}
\newcommand{\promptvar}[1]{\mbox{\color{blue!60!black}\normalfont\itshape\scriptsize$\langle$#1$\rangle$}}
\newcommand{\promptrole}[1]{\noindent{\small\textsf{\textbf{#1}}}\par\nobreak\vspace{1pt}}

\iclrfinalcopy 

\begin{document}
\newtheorem{definition}{Definition}
\newtheorem{theorem}{Theorem}
\newtheorem{corollary}{Corollary}

\maketitle
\begin{abstract}
Test-time scaling has enabled LLMs to solve increasingly complex reasoning, discovery, and optimization tasks.
However, existing approaches often gather observations while seeking evidence for current hypotheses, then generate new hypotheses conditioned on those hypotheses and their associated evidence.
This couples observation discovery to existing hypotheses, favoring local refinements over exploration of unknown hypotheses regions.
Inspired by the hypothetico-deductive method~\citep{hempel1945studies} and dual-space models of scientific discovery~\citep{klahr1988dual}, we introduce \system{}, a scientific reasoning harness that separately represents and jointly explores hypothesis and observation spaces.
By actively seeking new observations along with new hypotheses, \system{} can discover observations that its current hypotheses do not explain.
These unexplained observations expose gaps in its understanding, motivate new investigations, and guide the generation of hypotheses beyond refinements of existing ones.
Across diagnosis, abstract reasoning, and optimization benchmarks, \system{} outperforms evaluated baselines by up to 26.7 points.
Additional analysis shows that \system{} generates up to 4.5$\times$ more novel solutions.
\end{abstract}

\begin{figure}[!h]
    \centering
    \includegraphics[width=\linewidth]{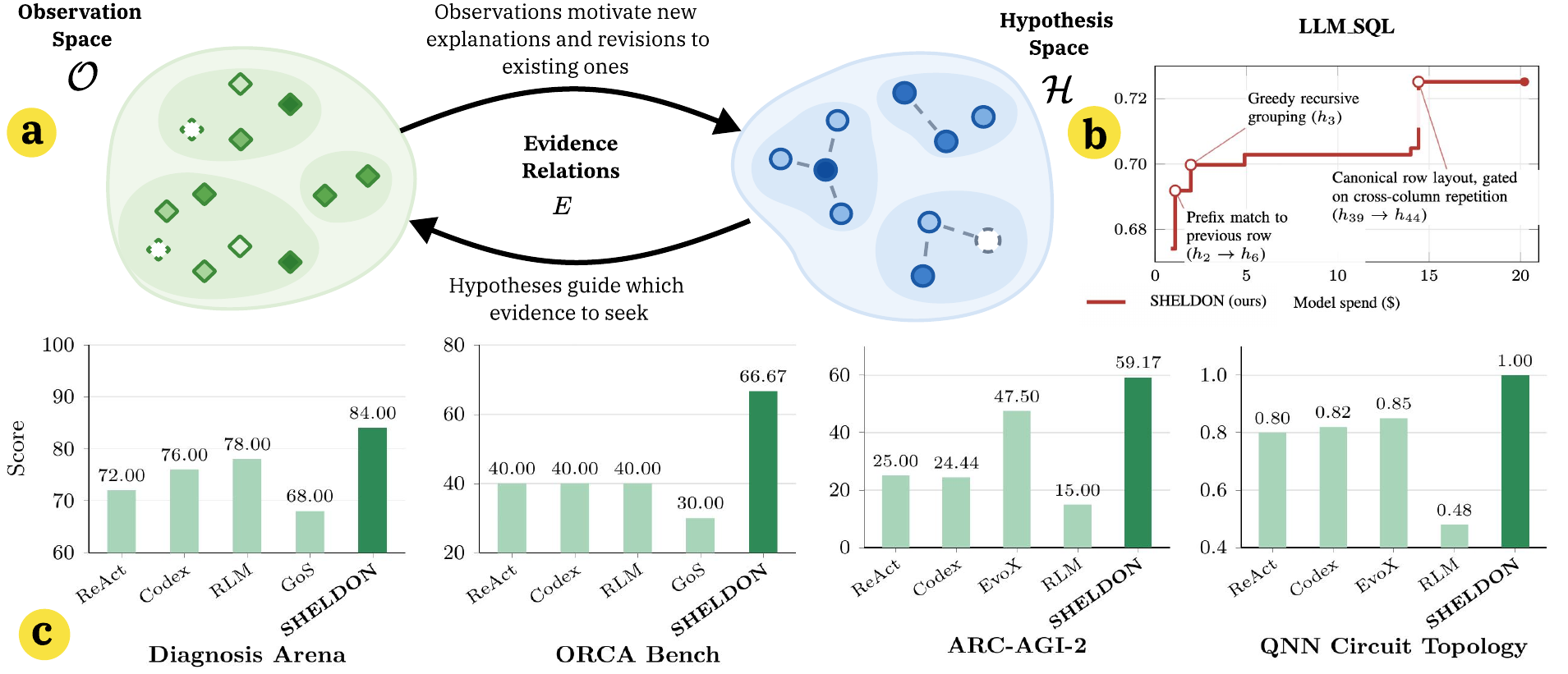}
    \vspace{-2mm}
    \caption{\logosystem{} jointly explores hypothesis and observation spaces, and outperforms strong baselines across tasks with and without explicit scoring functions. \textbf{(a)} Conceptual illustration of dual-space reasoning; edges represent refinement of hypotheses. \textbf{(b)} Improvements on a system optimization task. \textbf{(c)} Performance across diverse investigation tasks.}
    \label{fig:intro}
\end{figure}

\section{Introduction}

With advances in test-time scaling, LLMs have become increasingly capable of tackling complex tasks traditionally associated with general intelligence~\citep{snell2024scaling,venkatraman2025recursive,brown2024large}, including algorithm discovery~\citep{novikov2025alphaevolve, mitchener2025kosmos}, challenging mathematical reasoning~\citep{openai2026unitdistance}, and system optimization~\citep{cheng2025barbarians, cemri2026adaevolve}. 
For example, by using a swarm of 10000 concurrent agents, OpenAI resolves the Navier--Stokes Millennium Prize problem~\citep{openai2026navier}.

However, existing test-time scaling approaches face distinct limitations.
General coding agents such as Codex~\citep{codex} and Claude Code~\citep{claudecode} proceed sequentially, appending tool outputs to their context and using the resulting history to choose the next action.
This results in refinements to existing solution without ensuring that all accumulated evidence is reconciled to come up with the final solution.
Evolutionary methods~\citep{novikov2025alphaevolve, cemri2026adaevolve} broaden exploration by maintaining populations of candidate solutions.
Yet their search remains governed by predefined evolutionary mechanisms, and fitness-based selection only favors further exploration around discovered high-performing candidates.
EvoX~\citep{liu2026evox} introduces greater flexibility by searching over evolutionary strategies themselves.
Nevertheless, its exploration still remains anchored to the initial population and the candidates derived from it.

Recent specialized harnesses make the research state explicit: Arbor organizes hypotheses and experimental insights in a persistent tree~\citep{jin2026toward}, GoS maintains a graph of hypotheses and evidence~\citep{luo2026graph}, and ScientistOne explores parallel solution branches while preserving evidence provenance~\citep{meng2026scientistone}.
These structures enable later investigations to build on previous findings.
However, collecting evidence is guided by existing hypotheses which leads to observations that are already known or are refinements of existing hypotheses.
This leaves phenomena outside their scope unexamined.
Even comparing all current hypotheses may therefore overlook gaps in the assumptions they share.
Discovering genuinely different hypotheses and solutions requires opportunities to encounter observations that challenge shared assumptions

To broaden hypothesis exploration and strengthen investigation, we draw on the hypothetico-deductive method~\citep{hempel1945studies, hempel1966philosophy}, \citet{simon1974problem}'s work on rule induction, and its extension to scientific discovery through dual-space search~\citep{klahr1988dual}.
These perspectives characterize discovery as an iterative process in which hypotheses guide empirical investigation, and observations reshape subsequent hypotheses.
We adapt these ideas by separately representing and exploring \textbf{hypothesis} and \textbf{observation} spaces.
The hypothesis space supports \emph{exploitation} by retaining promising explanations and their investigation outcomes for further refinement.
The observation space supports \emph{exploration} by treating observations as explicit objects of discovery, including those that no current hypothesis adequately explains.
Such unexplained observations expose gaps in existing explanations, motivating new investigations and qualitatively different hypotheses.

We instantiate this process in \textbf{\logosystem{}} (Systematic Hypothesis Exploration and Learning through Directed Observation and Navigation), a scientific reasoning harness that maintains both spaces and connects them through evidence relations.
Through exploration, candidate-level investigation, comparative analysis, and exploitation, \system{} uses hypotheses to direct evidence collection and observations to assess competing explanations and identify new research directions.
The two spaces thus co-evolve, supporting refinement while reducing the risk of premature concentration on a narrow set of hypotheses.
As shown in Figure~\ref{fig:intro}, \system{} outperforms evaluated baselines across diagnosis, abstract reasoning, and optimization tasks by up to 26.7 points.
Notably, it discovers a Quantum Neural Network~\citep{liu2026skydiscover} topology achieving a perfect score of 1.00, surpassing both the strongest agent baseline (0.85) and the benchmark's hand-written reference topology (0.917).
Ablations demonstrate the contribution of the observation space, removing it reduces performance by 14.0 and 13.34 points on DiagnosisArena~\citep{zhu2026diagnosisarena} and ORCA-Bench~\citep{gong2026orca}, respectively.
Further analysis shows that, compared with GoS~\citep{luo2026graph}, \system{} increases the proportion of ground-truth causes discovered by 29.8 percentage points and the proportion of discovered causes investigated by 34.4 percentage points.
\section{Related Work}

\paragraph{Evolutionary Algorithms}
Recent work combines evolutionary search with LLMs to generate and refine populations of candidate solutions. FunSearch~\citep{romera2024mathematical} applies this paradigm to mathematical and algorithmic discovery, while AlphaEvolve~\citep{novikov2025alphaevolve} scales it to larger programs. Following works explored scientific equation discovery, code evolution, program optimization, and prompt optimization \citep{openevolve, shojaee2025llm, hemberg2024evolving, agrawal2026gepa}. More recent methods adapt the search process itself. ShinkaEvolve~\citep{lange2026shinkaevolve} improves sampling and model selection, AdaEvolve~\citep{cemri2026adaevolve} adjusts exploration and resource allocation, DeltaEvolve~\citep{jiang2026deltaevolve} tracks semantic deltas across candidate solutions, and EvoX~\citep{liu2026evox} jointly evolves solutions and search strategies. These methods primarily search around and evolve candidate solutions. \system{} instead also models observation spaces, allowing observations to motivate new hypotheses and discriminate among competing ones.

\paragraph{Research Agents}
LLM-based research agents increasingly automate scientific workflows spanning idea generation, literature review, experimentation, and synthesis \citep{lu2024ai, yamada2025ai, schmidgall2025agent, kon2025curie, swanson2025virtual}. CORAL~\citep{qu2026coral} supports long-running exploration with periodic interventions when progress stalls. Other work explicitly structures the research state: GoS~\citep{luo2026graph} represents hypotheses and evidence in a causal graph, but focuses investigation only on the highest-probability hypothesis. On the other hand, \system{} performs comparative analysis of hypotheses. ScientistOne~\citep{meng2026scientistone} links scientific claims to supporting sources and execution artifacts, and Arbor~\citep{jin2026toward} maintains hypotheses, artifacts, and evidence in a hypothesis tree. These systems largely organize evidence around existing hypotheses or artifacts. \system{} explicitly models the observation space, coupled to observation-driven hypothesis generation and targeted cross-hypothesis investigation.
Several evolutionary and research-agent methods, including AdaEvolve, EvoX, Arbor, and CORAL, are designed for settings with an explicit quantitative scoring function. Instead, \system{} can operate without such a scoring function, using observations as an implicit scoring function as shown on the medical diagnosis and system root-cause analysis benchmarks.

\section{Problem Formulation}
\begin{definition}[Investigative Problem]
An \textbf{investigative problem} $p$ is defined by a statement $x_p$, case material $\mathcal{M}_p$, a set of available tools $\mathcal{T}_p$,  
and a solution quality function $Q_p$, which may or may not be made available to the agent solving the problem. 
\end{definition}
The case material $\mathcal{M}_p$ comprises the problem-specific information provided for the investigation, such as system logs, seed programs, execution traces, or experimental measurements.
Tools $\mathcal{T}_p$ specify the capabilities available to the agent, such as code execution, database access, or simulators.
$Q_p$ may measure agreement with the ground truth, or performance under an external evaluation metric.

\begin{definition}[Investigation Task]
Given an investigative problem $p$ and a budget $B$, an LLM agent comprising a harness $\pi$ and a language model $m$ interacts sequentially with case material $\mathcal{M}_p$ and tools $\mathcal{T}_p$. At step $t$, the agent generates an action $a_t$ based on the interaction history $\tau_{t-1}=(a_1,o_1,\ldots,a_{t-1},o_{t-1}),$
and receives an observation $o_t$. Each step incurs cost $c_t$.
Let the solution quality function $\hat{Q}$ be $Q_p$ if it is available, and a generated estimate otherwise. 
After $T$ steps, the agent computes the final solution $\hat y_{\pi,m} = \Answer(\tau_{T})$ where the objective of \Answer is to maximize $\hat{Q}(\hat y_{\pi,m})$ subject to
$\sum_{t=1}^{T} c_t \le B$.
\end{definition}
\begin{figure}[!t]
    \centering
    \includegraphics[width=\linewidth]{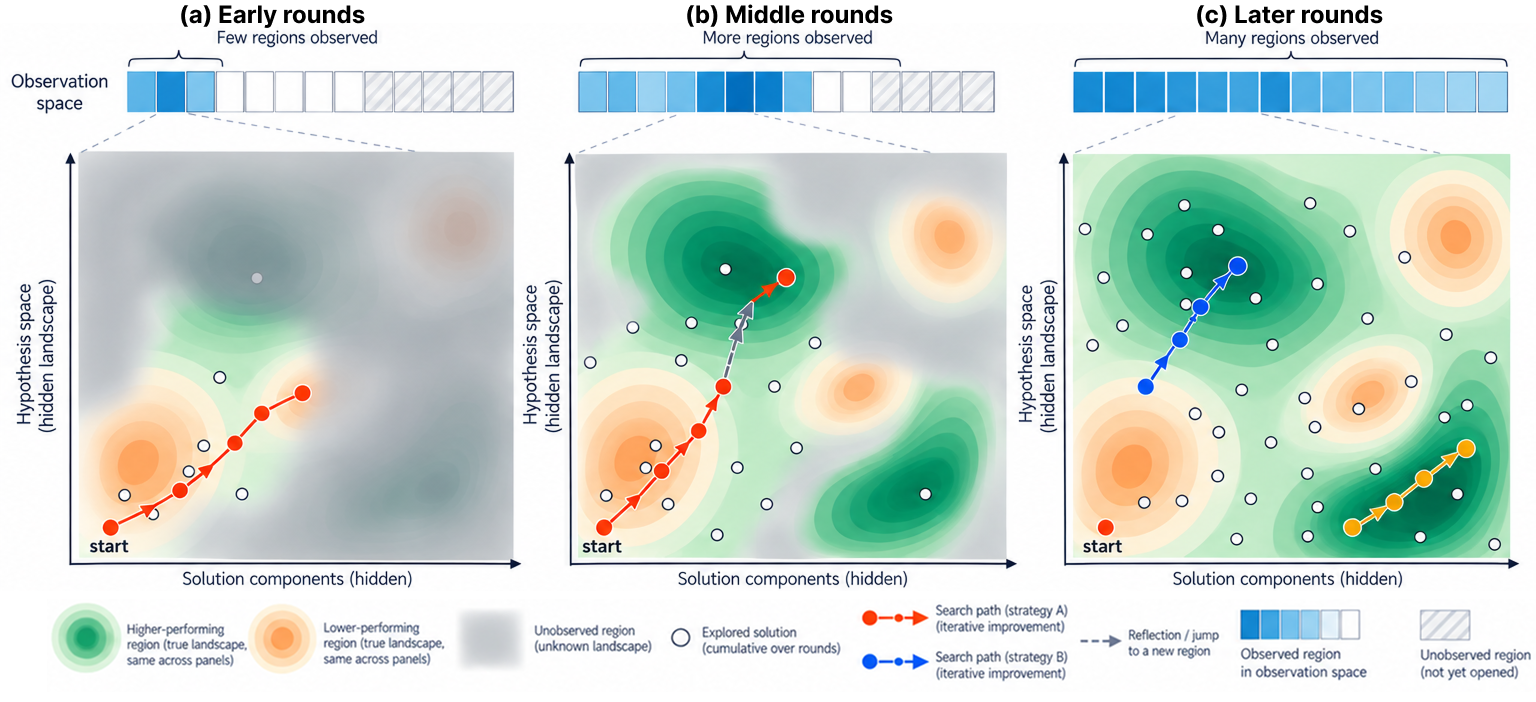}
    \caption{Conceptual illustration of evolution of the hypothesis space across successive rounds driven by discovery of newer observations. (a) Early rounds: Limited observational coverage constrains exploration to a small subset of the hypothesis space, producing potentially suboptimal solutions. (b) Middle rounds: New observations helps uncovers newer hypotheses regions guiding the search toward higher-performing solutions. (c) Later rounds: Broader observational coverage supports the exploration of multiple regions, allowing distinct solution strategies to emerge and achieve comparable performance. Blue bars represent observed regions, while hatched bars indicate unexplored regions. Green and orange contours denote higher- and lower-performing regions of the hypothesis space, respectively. Solid arrows indicate iterative solution improvement, and dashed arrows represent observation space-driven transitions. }
    \label{fig:concept_obs_hypo}
\end{figure}

\section{\logosystem{}: A Harness for Scientific Reasoning}
\label{sec:methodology}

\system{} organizes scientific reasoning as a coupled search over candidate hypotheses and observations acquired from the case material. Hypotheses guide which evidence to seek, while observations motivate new explanations and revisions to existing ones. Evolving maps of both spaces organize the alternatives under consideration and expose further directions for investigation, as shown in Figure~\ref{fig:concept_obs_hypo}.

\subsection{Dual-Space Representation}

\paragraph{Hypothesis and observation spaces.}
The dual-space account of \citet{klahr1988dual} describes scientific reasoning as coordinated search over a hypothesis space and an experiment space, with experiments providing evidence that distinguishes among hypotheses. We generalize the second space to an \textbf{observation space}, since evidence in our setting can arise from searches, inspections, program executions, explicit experiments, and other interactions with case material and tools.

We define the \textbf{hypothesis space} $H \in \mathcal{H}$ as the space of candidate explanations, solutions, or contributing ideas for a task. Depending on the task, a hypothesis may represent a proposed root cause, a diagnosis, a transformation rule, or a candidate design. To facilitate exploration of the hypotheses, we explicitly characterize the space with \textit{axes} and \textit{regions}, inspired by the literature on hypothesis scopes~\citep{thompson2023scope} and conceptual spaces~\citep{gardenfors2000conceptual}.
An axis identifies a dimension along which candidates may differ; its regions describe alternatives or portions of that dimension. Examples of axes include method families, underlying assumptions, or ways of determining free parameters. 
They provide an organization of the search space and a means of identifying qualitatively different investigative directions. 
At each time step, we track a \textit{working hypothesis set} $\wh_t=\{h_1,\ldots,h_n\} \subset H_t$, where $H_t$ is the hypothesis space at time $t$. 

Similarly, we define the \textbf{observation space} $O \in \mathcal{O}$ as the space of possible observations containing the evidence and evidence-grounded insights obtainable through interaction with the task environment. The observation space is also characterized by axes and regions, where possible axes include data sources, time ranges, entities or subsets, behavioral properties, scales, or tools. As observations also evolve iteratively, at each time step we track a working set of observations $\wo_t=\{o_1,\ldots,o_n\} \subset O_t$, where $O_t$ is the observation space at time $t$. 
It contains the cumulative observations currently used to guide investigation. These may describe individual findings or higher-level patterns composed from several findings. 

\begin{figure*}[t]
\centering

\begin{minipage}[t]{0.69\textwidth}
\vspace{0pt} 
\small
\vspace{2pt}
\hrule
\vspace{2pt}
\captionof{algorithm}{\system{} Dual-Space Reasoning Process}
\label{alg:sheldon}
\vspace{-4mm}
\hrule

\begin{algorithmic}[1]
\Require Investigative problem $p$, interaction budget $B$
\Ensure Final solution $\hat{y}$

\State Let the exploration space at time $t$ be
$S_t=\langle H_t,O_t,\wh_t,\wo_t,E_t\rangle$
\State $\hat{Q}\gets Q_p$ if one is available, and an estimated function otherwise.
\State $t\gets0$, $S_0\gets\textsc{Explore}(p,\emptyset)$

\While{budget remains and investigation has not stabilized}
    \ForAll{$h\in\wh_t$}
        \State $\wo_t,E_t\gets
        \UpdateObservations(\wo_t,\LocalTest(p,\hat{Q},S_t))$
    \EndFor
    \State $\wo_t,E_t\gets
    \UpdateObservations(\wo_t,\GlobalTest(p,\hat{Q},S_t))$
    \State $(\wh_t,\wo_t)\gets\revise(p,\hat{Q},S_t)$
    \State $S_{t+1}\gets\textsc{Explore}(p,S_t)$
    \State $t\gets t+1$
\EndWhile

\State $\hat{y}\gets\textsc{Answer}(p,\hat{Q},S_t)$
\State \Return $\hat{y}$
\end{algorithmic}
\hrule
\end{minipage}
\hfill
%
\begin{minipage}[t]{0.28\textwidth}
\vspace{0pt} 
\centering

\includegraphics[width=3.5cm]{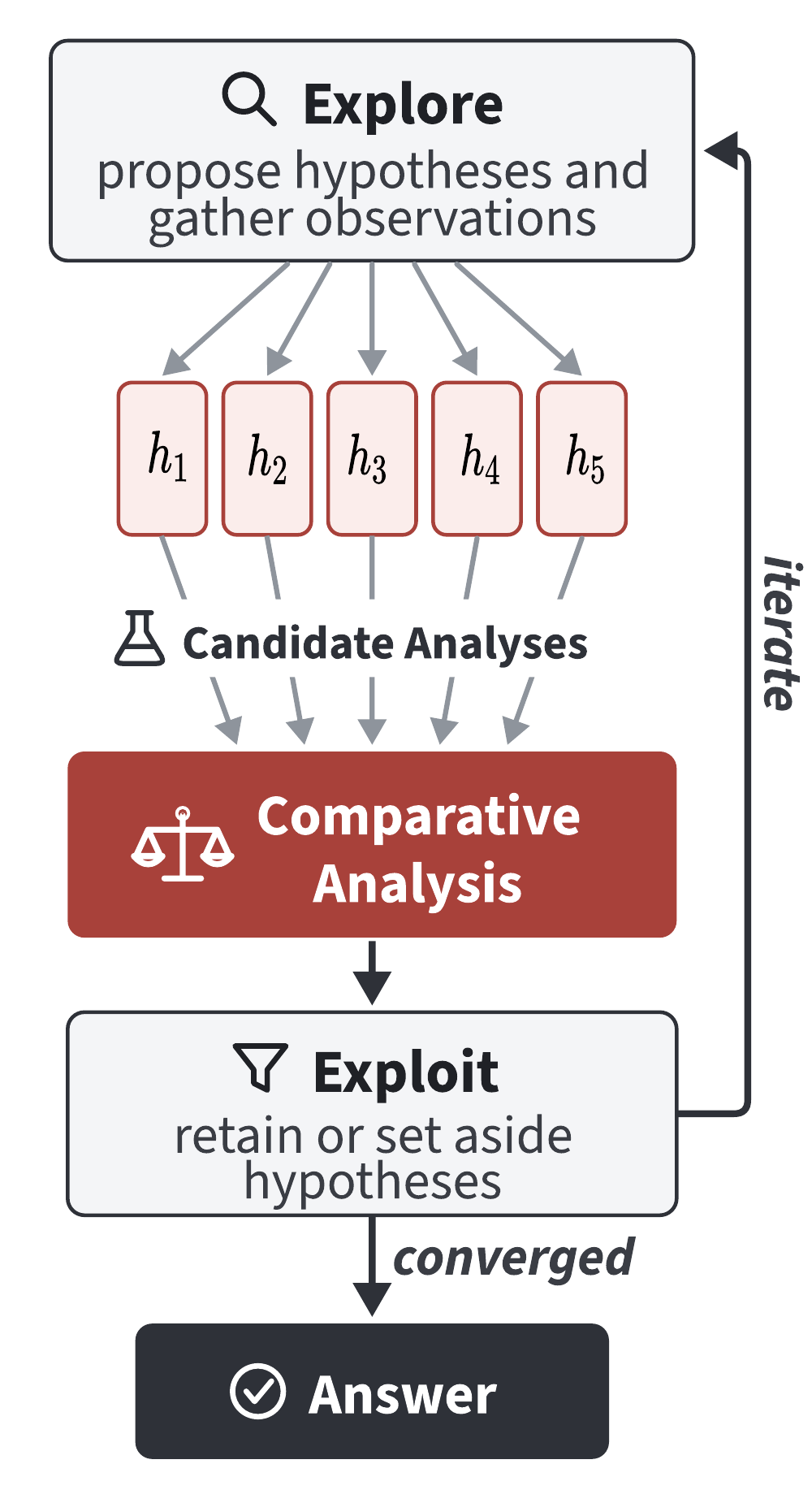}

\end{minipage}
\captionof{figure}{
\textbf{Overview of \system{}'s reasoning process.}
\system{} alternates between exploration, parallel candidate analysis,
comparative analysis across hypotheses, and exploitation.}
\label{fig:methodology2}
\end{figure*}

\subsection{Dual-Space Reasoning Process}
Algorithm~\ref{alg:sheldon} presents a high-level overview of the reasoning process.
Let $\hat{Q}$ be the solution quality function $Q_p$ if provided, and an estimate otherwise. 
The {\em state} of exploration at step $t$, $S_t = \langle H_t, O_t, \wh_t, \wo_t, E_t \rangle$, captures the cumulative result of the actions and observations up to time $t$: the current hypothesis and observation spaces ($H_t, O_t$), the working hypothesis and observation sets ($\wh_t, \wo_t$), and the evidence relations 
 \[ E_t: H \times O \rightarrow \{\textsc{Support},\textsc{Refute},\textsc{Neutral},\textsc{Unresolved}
 \}\] indicating the evidence status of an observation with respect to a hypothesis. Similar to Natural Language Inference (NLI)~\citep{condoravdi2003entailment,bowman2015large}, these categories describe whether an observation provides evidence for a hypothesis, provides evidence against it, is neutral with respect to it, or has a relationship that has not yet been assessed. Here, the relation captures a task-dependent evidential judgment rather than strict textual entailment or contradiction. For example, an observation may support a diagnostic explanation, expose a failure mode of a candidate design, or reveal the conditions under which an implementation succeeds. This provides the agent with a quantitative way of analyzing the working hypotheses given the working observation set.

To manage the working sets of hypotheses and observations, we define the
status of $h_i$, $\status(h_i)$ as 
one of \retained or \setaside, indicating whether it is actively considered or temporarily set aside based on the strength of its evidence. A set-aside hypothesis cannot be reconsidered until a cooldown period of a fixed number of rounds of investigation has elapsed (three rounds in our experiments; all hyperparameter values are listed in Table~\ref{tab:config-method}). 

We also introduce the boolean attribute \Provisional to each observation
because LLMs may occasionally make mistakes in their judgment~\citep{grand2026shoot} such as making wrong assumptions, buggy implementation, or being ambiguous. When a new observation is encountered, it is given a \Provisional status until it is encountered again. A \Provisional observation can be used to guide exploration and follow-up hypothesis testing, but it cannot be used to refute a hypothesis. 

We define $\Answer(p, \hat{Q}, S_t)$ as an agent-created function that computes a solution based on a set of hypotheses $\wh$ with respect to problem $p$ that maximizes the solution quality function $\hat{Q}$;
$\UpdateObservations(\wo, \wo')$ is a function that adds $\wo'$ to $\wo$. 

The investigation is an iterative process consisting of these steps, as shown in Figure~\ref{fig:methodology2}:
\begin{enumerate}
\item
\Explore performs actions to examine the case material, tools, and the existing exploration state. It defines new hypothesis and observation spaces, and proposes a new set of working hypotheses and observations.  

\item
\LocalTest gathers observations of individual hypotheses. 

\item 
\GlobalTest gathers observations of the full working hypothesis set as a whole, seeking evidence that helps compare the hypotheses. 

\item
\revise analyzes the gathered observations to refine the hypotheses, possibly dropping and adding hypotheses, and deciding their (\retained, \setaside) statuses.
\end{enumerate}

The process continues until the interaction budget $B$ is exhausted, the maximum number of rounds is reached (eight in all our experiments), or the investigation stabilizes, meaning that a round produces no new hypotheses, observations, or investigative directions. 
At termination, \system{} performs a final evidence assessment over the current working state and returns the candidate solution best supported by the accumulated evidence and available evaluations.

\begin{figure}[!t]
    \centering
    \includegraphics[width=0.9\linewidth]{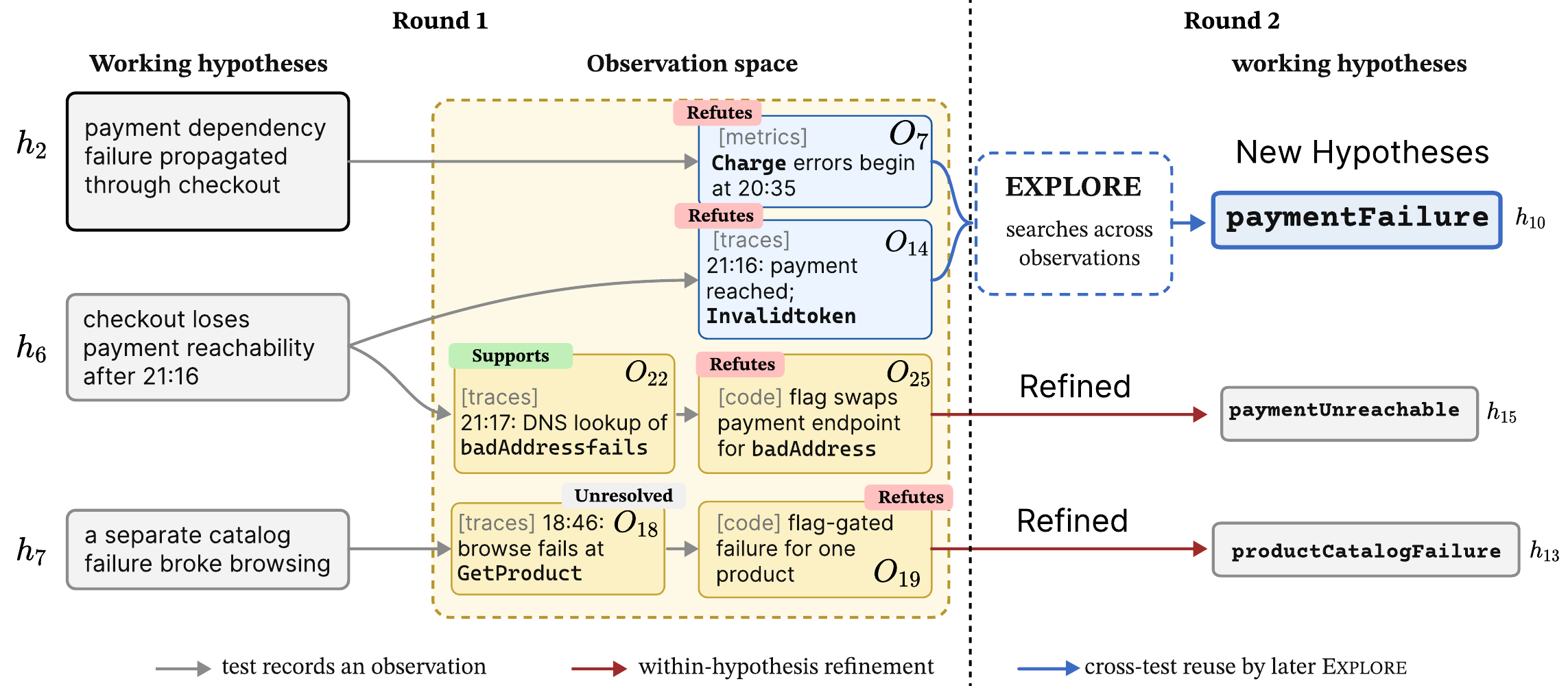}
    \caption{This ORCA-Bench incident contains concurrent catalog and payment failures.
In round 1, investigating hypothesis $h_2$ leads to the observation ($o_7$) that \textbf{\texttt{Charge}} errors
begin at 20:35, whereas $h_6$ leads to the observation ($o_{14}$) \textbf{\texttt{Invalid token}}.
In round 2, \textsc{Explore} combines these observations to form a new hypothesis ($h_{10}$) about \texttt{paymentFailure}, showing how the observation space leads to new hypotheses.
On the other hand, hypothesis ($h_{15}$) \texttt{paymentUnreachable} and
($h_{13}$) \texttt{productCatalogFailure} are obtained by refining hypotheses from round 1.}
    \label{fig:observations-space-example}
\end{figure}

\textbf{Explore.} The \Explore method performs actions to inspect the case material using available tools as well as the current exploration state to produce a collection of findings. From the findings, it constructs new hypothesis and observation spaces by defining (1) axes to help diversify hypothesis generation and (2) regions to track what has been exploited and what remains to be explored. It then creates a set of working hypotheses and observations. 
In the case where there are \setaside hypotheses, \Explore may also revisit hypotheses from the \emph{set-aside} pool once their cooldown period has elapsed. Although these hypotheses were previously deprioritized, their underlying ideas may still contribute useful components, assumptions, or directions for constructing new candidates. Reconsidering them therefore helps diversify the working set and reduces the risk of concentrating too narrowly on the currently dominant regions of the hypothesis space.

\textbf{Analyze Hypotheses.}
\system tests the hypotheses by first looking for evidence to support or refute individual candidates, then conducting a further analysis of the entire set of hypotheses. 

\LocalTest examines each working hypothesis individually in depth.
For each hypothesis, the agent investigates its assumptions and components, seeks supporting and refuting evidence, examines conditions under which it succeeds or fails, and follows up on findings that warrant further examination. Where a quality function $Q_p$ is available, it uses an LLM to implement and execute the function.
This process is repeated up to $N$ times ($N=3$ in our experiments).
After each round, the investigation inspects the resulting artifact, scores, or failures on particular cases. It updates the evidence relations between the hypothesis and the observations and returns all the new observations. \LocalTest thus develops an evidence-based account of a hypothesis's strengths, weaknesses, and failure modes.

 \GlobalTest considers the current working hypotheses jointly and seeks observations that distinguish among them. 
The agent identifies differences in candidates, assumptions, or predicted behaviors and chooses tests whose outcomes can discriminate among them. Such tests may favor some hypotheses while weakening others, expose failure modes shared across candidates, or reveal distinctions that were previously absent from the space maps. When scores are available, differences in candidate performance can motivate investigation into the source. It updates the evidence relations between the hypothesis and the observations and returns all the new observations. 
Together, \LocalTest and \GlobalTest enrich the working observation set and establish the
plausibility of individual hypotheses and discriminate between them.

\textbf{Exploit the Hypotheses.}
The \revise function then analyzes the hypotheses based on the observations. It can {\em refine} a hypothesis by improving it while preserving its core idea, or {\em repair} a hypothesis if its implemented artifact is defective. In the end, it classifies hypotheses as either \retained or \setaside to guide future exploration.

After revising the current hypotheses, the agent re-enters the exploration stage using the updated working hypothesis set, the working observation set, and the current hypothesis- and observation-space maps as well as the evidence relations, as shown in Figure~\ref{fig:observations-space-example}.
Exploration is conditioned on the structure already uncovered in both spaces, allowing the agent to balance exploitation of promising regions with exploration of underrepresented ones.
This enables the next generation of hypotheses to build on accumulated evidence without collapsing prematurely onto a narrow set of ideas.
Finally, \system selects $k$ hypotheses from the exploration and exploitation pools for further analysis in subsequent rounds ($k=5$ in our experiments).

\section{Experiments}
\label{sec:experiments}

\begin{table*}[t]
\centering
\caption{Details of benchmark tasks, case material, tools, how hypothesis and observation space could look like, and if the benchmark contains a Ground Truth (GT) or a Scoring Function (SF)}
\label{tab:benchmark-spaces}
\begingroup
\scriptsize
\setlength{\tabcolsep}{2.5pt}
\renewcommand{\arraystretch}{1.15}
\begin{tabularx}{\textwidth}{@{}
  Y{1.25}
  Y{0.85}
  Y{0.80}
  Y{1.35}
  Y{1.35}
  Y{0.40}
@{}}
\toprule
\textbf{Benchmark} & \textbf{Case material} & \textbf{Tools} & \textbf{Hypothesis space }& \textbf{Observation space} & \textbf{Metric} \\
\midrule
ORCA Bench: Find the cause of a service fault
& Telemetry, service code, incident report
& Container shell; Grafana API
& Fault explanations: service, fault type, and failure mechanism
& Metrics, logs, traces, code, and incident timeline
& GT \\
\addlinespace
Diagnosis-Arena: Diagnose a patient
& Patient record: history, exam, tests, and notes
& EviDx record readers
& Candidate diagnoses and their predicted symptoms and test results
& History, labs, imaging, pathology, and treatment response
& GT \\
\addlinespace
ARC-AGI-2: Infer a grid transformation rule
& Example grid pairs, test grid, seed program
& Python sandbox
& Candidate rules: symmetry, movement, replacement, encoding, and tiling
& Grid changes, object shapes, patterns across examples, and synthetic tests
& GT \\
\addlinespace
QNN topology: Optimise a two-qubit circuit
& Circuit seed and training harness
& Python sandbox; evaluator
& Circuit design explanations: gate count, connections, rotation axes, and parameters
& Circuit scores, gate-removal tests, design variants, and repeated runs
& SF \\
\addlinespace
ADRS: Optimise a systems program
& Program seed, evaluator, workload
& Python sandbox; evaluator
& Algorithm explanations: greedy methods, grouping, allocation, search, and scheduling
& Program scores, validity checks, test workloads, edge cases, and timing
& SF \\
\addlinespace
DOCK-STRING: Design a drug-like molecule
& Target and scored molecule library
& Docking, property, and similarity tools; Python
& Molecule design explanations: scaffold, chemical groups, size, and rigidity
& Docking scores, molecule properties, similar molecules, and chemical variants
& SF \\
\bottomrule
\end{tabularx}
\endgroup
\end{table*}

\textbf{Benchmarks.} We evaluate \system{} across 10 tasks with and without explicit quality scoring functions.
Tasks without a scoring function are particularly challenging because the agent must infer whether a candidate solution is plausible from the available evidence. 
For this setting, we use DiagnosisArena~\citep{zhu2026diagnosisarena}, a differential-diagnosis benchmark, ORCA-Bench~\citep{gong2026orca}, a system-level root-cause analysis benchmark, and ARC-AGI-2~\citep{chollet2025arc}, where the agent must infer the transformation rule from input-output pairs. 
We also evaluate \system{} on tasks with an explicit scoring function, including molecular optimization~\citep{garcia2022dockstring}, quantum neural network topology optimization, and five additional ADRS system-optimization tasks~\citep{liu2026skydiscover}. 
We summarize these benchmarks in Table~\ref{tab:benchmark-spaces}, along with what potential hypothesis and observation space could have for \system{}.
For all benchmarks, we use the official evaluation metrics and scoring scripts.
We provide more details on benchmark tasks and evaluation metrics in Appendix~\ref{appendix:benchmarks}.

\textbf{Baselines.}
We evaluate several baselines and use GPT-5.6-terra with high reasoning effort for all. \emph{Iterative coding agent:} Codex~\citep{codex} using \texttt{codex-cli} v0.128.0. ReAct~\citep{yao2022react}, an agent that interleaves reasoning with tool-based actions.
\emph{Evolutionary:} EvoX~\citep{liu2026evox} requires an initial solution set and a quality scoring function.
\emph{Research agents:} ScientistOne~\citep{meng2026scientistone} and GoS~\citep{luo2026graph}.
\emph{Autonomous agents:} Arbor~\citep{jin2026toward} and CORAL~\citep{qu2026coral}, both require an initial artifact and a quality scoring function. RLM~\citep{zhang2025recursive}, a recursive reasoning agent.
Since the ScientistOne framework is not publicly available, we evaluate its released ADRS programs.
Since system optimization tasks can vary from hardware to hardware, we evaluate all methods on the same hardware. \system{} uses GPT-5.6-terra for all reasoning stages and delegates code implementation to GPT-5.6-luna.

\subsection{Performance on diverse benchmarks}

\textbf{\system{} performs across diverse reasoning and discovery tasks.} It achieves the highest reported score on all four tasks, outperforming baselines across reasoning agents, coding agents, recursive agents, and meta-evolution, as shown in Table~\ref{tab:diagnostic-reasoning-results}.
The strongest baseline varies across domains: RLM leads on DiagnosisArena, while EvoX leads on ARC-AGI-2 and QNN Circuit Topology, but \system{} maintains its advantage throughout, showing its generalization capability.

\textbf{\system{} achieves substantial gains on tasks that require distinguishing competing explanations from evidence.} DiagnosisArena evaluates clinical diagnostic reasoning, while ORCA-Bench requires identifying incident root causes from system logs, metrics, traces, and source code. \system{} achieves 84.00 on DiagnosisArena, exceeding RLM's 78.00 by 6 percentage points and Codex's 76.00 by 8 points. On ORCA-Bench, it reaches 66.67, improving over the strongest reported baseline score of 40.00 by 26.67 points. 
We believe these gains come from \system{}'s rigor in explicitly comparing and analyzing different hypotheses.

\textbf{\system{} also advances abstract reasoning and circuit discovery through structured exploration of candidate solutions.} ARC-AGI-2 tests the discovery and application of transformation rules, including interacting and context-dependent rules.
QNN Circuit Topology evaluates circuit structures for a classification task.
On ARC-AGI-2 under \$2.5 budget per task, \system{} achieves 59.17, exceeding EvoX's 47.50 by 11.67 percentage points. In Figure~\ref{fig:arc-example}, we show how \system{} uses hypotheses to come up with new observations and how observations then influence the newer hypotheses.
On QNN Circuit Topology, it reaches a perfect score of 1.00. 
\system{} discovers a ten-gate, two-qubit classifier that combines parameterized rotations and CNOT gates, followed by a Hadamard gate on the second qubit, with the readout taken as $\langle Z\rangle$ on the first qubit, to classify angle-encoded inputs (Figure~\ref{fig:qnn-topology} in Appendix~\ref{appendix:benchmarks}).
These results support the effectiveness of maintaining alternative hypotheses and refining them through investigation across both rule induction and executable design, extending \system{}'s advantages beyond explanatory diagnosis.

\begin{table*}[t]
\centering
\caption{
Performance on diagnostic and reasoning benchmarks across different
agent and optimization paradigms. Cell shading indicates the top three
performing methods within each benchmark, with darker shading denoting
higher rank. EvoX requires a scoring function, and GoS requires benchmark-specific sub-agent prompts.
}
\label{tab:diagnostic-reasoning-results}

\setlength{\tabcolsep}{3.2pt}
\renewcommand{\arraystretch}{1.05}

\resizebox{0.80\textwidth}{!}{%
\begin{tabular}{llcccc}
    \toprule
    \textbf{Method}
    & \textbf{Method Family}
    & \makecell{\textbf{Diagnosis}\\\textbf{Arena}}
    & \makecell{\textbf{ORCA}\\\textbf{Bench}}
    & \makecell{\textbf{ARC}\\\textbf{AGI-2}}
    & \makecell{\textbf{QNN Circuit}\\\textbf{Topology}} \\
    
    \midrule

    ReAct~\citep{yao2022react}
    & Reasoning agent
    & \hspace{2pt}72.00\hspace{2pt}
    & \heatcell{20}{\hspace{2pt}40.00\hspace{2pt}}
    & \heatcell{12}{\hspace{2pt}25.00\hspace{2pt}}
    & \hspace{2pt}0.80\hspace{2pt}
    \\

    Codex~\citep{codex}
    & Coding agent
    & \heatcell{12}{\hspace{2pt}76.00\hspace{2pt}}
    & \heatcell{20}{\hspace{2pt}40.00\hspace{2pt}}
    & \hspace{2pt}24.44\hspace{2pt}
    & \heatcell{12}{\hspace{2pt}0.82\hspace{2pt}}
    \\

    EvoX~\citep{liu2026evox}
    & Meta-evolution
    & $\times$
    & $\times$
    & \heatcell{24}{\hspace{2pt}47.50\hspace{2pt}}
    & \heatcell{24}{\hspace{2pt}0.85\hspace{2pt}}
    \\

    RLM~\citep{zhang2025recursive}
    & Recursive agent
    & \heatcell{24}{\hspace{2pt}78.00\hspace{2pt}}
    & \heatcell{20}{\hspace{2pt}40.00\hspace{2pt}}
    & \hspace{2pt}15.00\hspace{2pt}
    & \hspace{2pt}0.48\hspace{2pt}
    \\

    GoS~\citep{luo2026graph}
    & Reasoning agent
    & \hspace{2pt}68.00\hspace{2pt}
    & \hspace{2pt}30.00\hspace{2pt}
    & $\times$
    & $\times$
    \\

    \textbf{\logosystem{} (Ours)}
    & Scientific Reasoning
    & \heatcell{45}{\hspace{2pt}\textbf{84.00}\hspace{2pt}}
    & \heatcell{45}{\hspace{2pt}\textbf{66.67}\hspace{2pt}}
    & \heatcell{45}{\hspace{2pt}\textbf{59.17}\hspace{2pt}}
    & \heatcell{45}{\hspace{2pt}\textbf{1.00}\hspace{2pt}}
    \\

    \bottomrule
\end{tabular}%
}

\end{table*}

\begin{table}[t]
\centering
\caption{Molecular optimization. Yellow cells beat the best molecule in the DOCKSTRING library.}
\label{tab:docking_results}
\renewcommand{\arraystretch}{1.0}
\setlength{\tabcolsep}{2pt}
\resizebox{0.62\textwidth}{!}{%
\begin{tabular}{lrrrrrr}
\toprule
\textbf{System}
& \textbf{ADRB1}
& \textbf{CDK2}
& \textbf{PPARA}
& \textbf{PPARG}
& \textbf{F2}
& \textbf{PGR} \\
\midrule

Codex
& \heatcell{8}{5.97}
& \heatcell{8}{6.92}
& \heatcell{8}{7.46}
& \heatcell{8}{6.38}
& \heatcell{40}{9.31}
& \heatcell{8}{7.40} \\

ReAct
& \heatcell{16}{8.49}
& \heatcell{28}{8.32}
& \heatcell{28}{8.98}
& \heatcell{28}{8.40}
& \heatcell{8}{7.79}
& \heatcell{16}{7.67} \\

EvoX
& \heatcell{28}{8.63}
& \heatcell{16}{7.19}
& \heatcell{16}{8.62}
& \heatcell{16}{7.41}
& \heatcell{16}{8.36}
& \heatcell{28}{9.17} \\

CORAL
& \heatcell{40}{11.01}
& \heatcell{40}{9.32}
& \heatcell{40}{9.64}
& \heatcell{55}{\highlight{\textbf{9.69}}}
& \heatcell{28}{9.10}
& \heatcell{55}{\highlight{\textbf{10.65}}} \\

\textbf{\logosystem{}}
& \heatcell{55}{\textbf{11.26}}
& \heatcell{55}{\highlight{\textbf{10.26}}}
& \heatcell{55}{\textbf{9.85}}
& \heatcell{40}{9.47}
& \heatcell{55}{\highlight{\textbf{9.85}}}
& \heatcell{40}{\highlight{9.99}} \\
\midrule
\textit{Library best}
& 11.33 & 9.53 & 9.91 & 9.53 & 9.31 & 9.74 \\
\bottomrule
\end{tabular}}
\end{table}

\textbf{\system{} achieves the best performance on four of six molecular targets and exceeds the library-best reference on three.} For the molecular optimization benchmark, \system{} outperforms Codex, ReAct, and EvoX on all six targets and is competitive with CORAL, achieving the best score on ADRB1, CDK2, PPARA, and F2.
On PPARG and PGR, CORAL obtains the highest scores of 9.69 and 10.65, respectively, while \system{} reaches 9.47 and 9.99. 
CORAL extensively uses search to retrieve existing molecules from the database and their properties, and retain the information in context. However, we observe that it exhausts the token budget during longer runs.
Notably, \system{} exceeds the library-best reference scores on CDK2, F2, and PGR, demonstrating that its observation space-guided hypothesis exploration and revision can lead to novel solutions.

\subsection{System Optimization Tasks}
\textbf{\system{} delivers its clearest gains where further investigation has room to improve solution quality.} Table~\ref{table:adrs} reports results over three runs per method, with \system{} achieving the best or tied-best mean and best-run performance across all five tasks.
Its strongest advantages emerge on transaction scheduling and LLM\_SQL, where inspection of the discovered programs reveals distinct strategies for improving solution quality.
For transaction scheduling, \system{} evaluates transaction-position pairs during construction, then refines the schedule through deterministic swap and relocation searches.
This construction-intensive strategy achieves the best mean and ``best'' score, exceeding all baselines.
For LLM\_SQL, \system{} measures repetition patterns and selects between recursive partitioning and per-row reordering. 
Its implementation runs in 4.3s compared with EvoX's 7.1s, achieving a better combined cache-reuse and runtime objective.
This illustrates that \system's general hypothetico-deductive scientific reasoning is capable of deriving domain-specific strategies to result in improvements in both average and peak performance.

\textbf{\system{} exhibits robust performance across benchmarks, including tasks with limited observed headroom.} Its best or tied-best means across all five tasks are consistent with its emphasis on maintaining alternatives and repeatedly reassessing them against evidence. Cloudcast illustrates the distinction between finding a strong solution and doing so consistently: Codex achieves a best transfer cost of 618.10 but a mean of 677.10, whereas \system{} achieves a best of 618.07 and a mean of 618.09. This consistency remains informative where best-run scores show apparent saturation. 
The best performance for EPLB, PRISM, and Cloudcast is mostly the same across all methods, suggesting that they are relatively simple problems that can be handled with weaker techniques.

\begin{table*}[t]
\centering
\caption{Performance across ADRS system-level optimization task benchmarks.
``Mean'' reports the average score over 3 runs. ``Cost to best'' reports the cost in USD required to obtain the
best-performing solution; lower is better. The darkest cell marks the best result. ``Human best'' is the best human-designed method as reported by \citet{liu2026evox}, measured on their hardware.}
\label{table:adrs}
\small
\setlength{\tabcolsep}{3.2pt}
\renewcommand{\arraystretch}{1.15}

\resizebox{\textwidth}{!}{%
\begin{tabular}{l
    ccc
    ccc
    ccc
    ccc
    ccc}
    \toprule
    & \multicolumn{3}{c}{\textbf{EPLB}}
    & \multicolumn{3}{c}{\textbf{PRISM}}
    & \multicolumn{3}{c}{\textbf{Cloudcast}}
    & \multicolumn{3}{c}{\textbf{Txn Scheduling}}
    & \multicolumn{3}{c}{\textbf{LLM\_SQL}} \\

    \cmidrule(lr){2-4}
    \cmidrule(lr){5-7}
    \cmidrule(lr){8-10}
    \cmidrule(lr){11-13}
    \cmidrule(lr){14-16}

    \textbf{System}
    & \makecell{\textbf{Mean}}
    & \textbf{Best}
    & \makecell{\textbf{Cost to}\\\textbf{best} $\downarrow$}

    & \makecell{\textbf{Mean}}
    & \textbf{Best}
    & \makecell{\textbf{Cost to}\\\textbf{best} $\downarrow$}

    & \makecell{\textbf{Mean}} $\downarrow$
    & \textbf{Best} $\downarrow$
    & \makecell{\textbf{Cost to}\\\textbf{best} $\downarrow$}

    & \makecell{\textbf{Mean}}
    & \textbf{Best}
    & \makecell{\textbf{Cost to}\\\textbf{best} $\downarrow$}

    & \makecell{\textbf{Mean}}
    & \textbf{Best}
    & \makecell{\textbf{Cost to}\\\textbf{best} $\downarrow$} \\
    \midrule
    \textit{Human best}
    & -- & 0.1265 & --
    & -- & 21.89 & --
    & -- & 626.24 & --
    & -- & 2724.8 & --
    & -- & 0.6920 & -- \\
    \midrule

    Codex
    & \heatcell{55}{\textbf{0.1441}} & \heatcell{55}{\textbf{0.1441}} & 0.45
    & \heatcell{8}{25.60} & \heatcell{28}{25.65} & 0.23
    & \heatcell{0}{677.10} & \heatcell{16}{618.10} & 0.24
    & \heatcell{0}{3069} & \heatcell{0}{3558} & 0.19
    & \heatcell{0}{0.6877} & \heatcell{0}{0.7109} & 0.22 \\

    ReAct
    & \heatcell{28}{0.1440} & \heatcell{28}{0.1440} & 0.28
    & \heatcell{16}{26.24} & \heatcell{55}{\textbf{26.26}} & 0.54
    & \heatcell{3}{621.11} & \heatcell{16}{618.10} & 0.49
    & \heatcell{8}{3932} & \heatcell{8}{4000} & 0.30
    & \heatcell{0}{0.7001} & \heatcell{0}{0.7069} & 0.64 \\

    \midrule

    EvoX 
    & \heatcell{0}{0.1318} & \heatcell{3}{0.1433} & 9.70
    & \heatcell{55}{\textbf{26.26}} & \heatcell{55}{\textbf{26.26}} & 2.00
    & \heatcell{28}{618.10} & \heatcell{28}{618.08} & 0.15
    & \heatcell{16}{4009} & \heatcell{16}{4184} & 9.51
    & \heatcell{28}{0.7149} & \heatcell{16}{0.7196} & 12.58 \\

    ScientistOne
    & \heatcell{3}{0.1319} & \heatcell{28}{0.1440} & --
    & \heatcell{55}{\textbf{26.26}} & \heatcell{55}{\textbf{26.26}} & --
    & \heatcell{0}{624.10} & \heatcell{16}{618.10} & --
    & \heatcell{0}{2533} & \heatcell{3}{3962} & --
    & \heatcell{3}{0.7034} & \heatcell{3}{0.7137} & -- \\

    Arbor
    & \heatcell{8}{0.1433} & \heatcell{8}{0.1435} & 6.86
    & \heatcell{28}{26.25} & \heatcell{55}{\textbf{26.26}} & 2.31
    & \heatcell{16}{618.80} & \heatcell{16}{618.10} & 2.52
    & \heatcell{3}{3668} & \heatcell{0}{3759} & 4.79
    & \heatcell{16}{0.7130} & \heatcell{28}{0.7218} & 1.23 \\

    CORAL
    & \heatcell{16}{0.1436} & \heatcell{16}{0.1436} & 12.01
    & \heatcell{55}{\textbf{26.26}} & \heatcell{55}{\textbf{26.26}} & 0.30
    & \heatcell{8}{619.00} & \heatcell{8}{619.00} & 0.24
    & \heatcell{28}{4077} & \heatcell{28}{4201} & 10.89
    & \heatcell{8}{0.7115} & \heatcell{8}{0.7163} & 1.23 \\

    \midrule

    \textbf{\logosystem{}}
    & \heatcell{55}{\textbf{0.1441}} & \heatcell{55}{\textbf{0.1441}} & 0.80
    & \heatcell{55}{\textbf{26.26}} & \heatcell{55}{\textbf{26.26}} & 1.28
    & \heatcell{55}{\textbf{618.09}} & \heatcell{55}{\textbf{618.07}} & 6.84
    & \heatcell{55}{\textbf{4198}} & \heatcell{55}{\textbf{4274}} & 9.84
    & \heatcell{55}{\textbf{0.7204}} & \heatcell{55}{\textbf{0.7245}} & 14.42 \\
    \bottomrule
\end{tabular}%
}
\end{table*}

\section{Analysis}

\textbf{Each component of \system{} plays a distinct role in sustaining effective investigation.}
We ablate comparative analysis, the observation space, and exploration from the second round onward, with the latter allowing the system to continue refining existing hypotheses without introducing new directions (Table~\ref{tab:ablations}). 
All three components contribute to performance, with the observation space and comparative analysis producing the most consistent degradation across all benchmarks when removed.
Continued exploration is also important, particularly on ORCA-Bench and the QNN topology task, where removing it causes a substantial drop.
Comparative analysis further improves results by explicitly contrasting competing hypotheses and actively seeking evidence that can distinguish between them.

\begin{table}[t]
    \centering
    \caption{
        \system{} ablation.
        \textbf{Bold} denotes the change relative to the full \system{} configuration.
    }
    \label{tab:ablations}
    \tiny
    \setlength{\tabcolsep}{8pt}
    \renewcommand{\arraystretch}{1.0}
\resizebox{0.92\textwidth}{!}{%
    \begin{tabular}{lcccc}
        \toprule
        \textbf{Variant}
        & \textbf{DiagnosisArena} $\uparrow$
        & \textbf{ORCA-Bench} $\uparrow$
        & \textbf{ARC-AGI-2} $\uparrow$
        & \textbf{QNN Topology} $\uparrow$\\
        \midrule

        \textbf{\logosystem{}} (full)
        & \textbf{84.00}
        & \textbf{66.67} 
        & \textbf{59.17} 
        & \textbf{1.00} \\

        \quad w/o comparative analysis
        & 78.00 {(\textbf{-6.00})}
        & 56.67 {(\textbf{-10.00})}
        & 50.00 {(\textbf{-9.17})}
        & 0.8833 {(\textbf{-0.117})}\\

        \quad w/o observation space
        & 70.00 { (\textbf{-14.00})}
        & 53.33 { (\textbf{-13.34})}
        & 50.00 {(\textbf{-9.17})}
        & 0.8667 {(\textbf{-0.133})}
        \\

        \quad w/o continued exploration
        & 82.00 { (\textbf{-2.00})}
        & 53.33 { (\textbf{-13.34})}
        & 57.70 {(\textbf{-1.47})}
        & 0.7333 {(\textbf{-0.267})}\\
        \bottomrule

    \end{tabular}}
\end{table}

\begin{figure}[!t]
    \centering
    \resizebox{\textwidth}{!}{%
    \begin{tikzpicture}
\begin{groupplot}[
  group style={group size=5 by 1, horizontal sep=1.2cm},
  width=3.5cm, height=3.9cm,
  xlabel={model spend (\$)}, xmin=0,
  xmajorgrids, ymajorgrids, grid style={gray!20},
  tick label style={font=\scriptsize}, label style={font=\scriptsize},
  title style={font=\bfseries\small, text height=1.6ex, text depth=0.3ex},
  scaled y ticks=false,
  yticklabel style={/pgf/number format/fixed, /pgf/number format/precision=4},
  legend style={font=\scriptsize, draw=none, legend columns=4, column sep=8pt},
]
\nextgroupplot[title={EPLB}, ymin=0.12, ymax=0.146, xmax=20.0, ylabel={best score}]
\addplot[const plot, arbor, line width=1.0pt, densely dashed] coordinates {(0.079,0.125657) (0.499,0.12605) (0.998,0.126052) (2.527,0.126053) (3.242,0.143481) (4.616,0.143483) (4.616,0.143483)};
\addplot[only marks, mark=*, mark size=1.8pt, arbor, forget plot] coordinates {(4.616,0.143483)};
\addplot[const plot, coral, line width=1.0pt] coordinates {(0.147,0.125689) (0.231,0.142963) (0.658,0.143574) (1.476,0.143584) (1.772,0.1436) (4.040,0.143623) (8.226,0.143623) (10.348,0.143627) (12.013,0.143631) (19.232,0.143631)};
\addplot[only marks, mark=*, mark size=1.8pt, coral, forget plot] coordinates {(19.232,0.143631)};
\addplot[const plot, evox, line width=1.0pt] coordinates {(0.000,0.125654) (0.054,0.125891) (0.103,0.12598) (0.150,0.12602) (0.253,0.12602) (0.351,0.126036) (1.346,0.126045) (1.649,0.126053) (1.754,0.1261) (4.446,0.143181) (4.490,0.143239) (4.595,0.143245) (4.643,0.143263) (4.804,0.1433) (5.644,0.143359) (5.691,0.143373) (5.850,0.1434) (6.121,0.1434)};
\addplot[only marks, mark=*, mark size=1.8pt, evox, forget plot] coordinates {(6.121,0.1434)};
\addplot[const plot, ours, line width=1.6pt] coordinates {(0.094,0.125663) (0.204,0.142946) (0.460,0.142948) (0.799,0.143495) (3.961,0.143522) (4.911,0.143522)};
\addplot[only marks, mark=*, mark size=1.8pt, ours, forget plot] coordinates {(4.911,0.143522)};
\nextgroupplot[title={PRISM}, ymin=26.0, ymax=26.3, xmax=19.2]
\addplot[const plot, arbor, line width=1.0pt, densely dashed] coordinates {(0.068,26) (0.563,26) (0.881,26) (2.441,26) (3.293,26.0652) (3.410,26.1867) (3.896,26.2285) (5.158,26.2285)};
\addplot[only marks, mark=*, mark size=1.8pt, arbor, forget plot] coordinates {(5.158,26.2285)};
\addplot[const plot, coral, line width=1.0pt] coordinates {(0.200,26.212) (0.297,26.256) (18.437,26.256)};
\addplot[only marks, mark=*, mark size=1.8pt, coral, forget plot] coordinates {(18.437,26.256)};
\addplot[const plot, evox, line width=1.0pt] coordinates {(0.000,26) (0.054,26) (0.115,26) (0.131,26) (0.164,26) (0.604,26.1246) (0.889,26.2035) (0.918,26.2035) (1.939,26.2271) (3.223,26.256) (3.223,26.256) (4.284,26.256)};
\addplot[only marks, mark=*, mark size=1.8pt, evox, forget plot] coordinates {(4.284,26.256)};
\addplot[const plot, ours, line width=1.6pt] coordinates {(0.053,26) (0.158,26) (0.512,26) (0.548,26) (0.611,26) (1.278,26.256) (4.745,26.256)};
\addplot[only marks, mark=*, mark size=1.8pt, ours, forget plot] coordinates {(4.745,26.256)};
\nextgroupplot[title={Cloudcast}, ymin=615, ymax=640, xmax=12.7, y dir=reverse, ylabel={total cost ($\downarrow$)}, legend to name=costlegend]
\addplot[const plot, arbor, line width=1.0pt, densely dashed] coordinates {(0.071,640) (0.675,640) (1.807,633.127) (2.387,627.114) (2.856,618.085) (3.118,618.085)};
\addplot[only marks, mark=*, mark size=1.8pt, arbor, forget plot] coordinates {(3.118,618.085)};
\addlegendentry{Arbor}
\addplot[const plot, coral, line width=1.0pt] coordinates {(0.162,640) (0.243,618.103) (12.252,618.103)};
\addplot[only marks, mark=*, mark size=1.8pt, coral, forget plot] coordinates {(12.252,618.103)};
\addlegendentry{CORAL}
\addplot[const plot, evox, line width=1.0pt] coordinates {(0.000,640) (0.048,640) (0.116,633.148) (0.204,618.103) (6.601,618.103)};
\addplot[only marks, mark=*, mark size=1.8pt, evox, forget plot] coordinates {(6.601,618.103)};
\addlegendentry{EvoX}
\addplot[const plot, ours, line width=1.6pt] coordinates {(0.629,640) (1.373,627.122) (1.596,618.103) (6.837,618.07) (10.246,618.07)};
\addplot[only marks, mark=*, mark size=1.8pt, ours, forget plot] coordinates {(10.246,618.07)};
\addlegendentry{SHELDON (ours)}
\nextgroupplot[title={Txn Scheduling}, ymin=3500, ymax=4300, xmax=15]
\addplot[const plot, arbor, line width=1.0pt, densely dashed] coordinates {(0.071,3500) (0.787,3500) (2.278,3500) (4.060,3500) (4.731,3623.19) (5.752,3623.19)};
\addplot[only marks, mark=*, mark size=1.8pt, arbor, forget plot] coordinates {(5.752,3623.19)};
\addplot[const plot, coral, line width=1.0pt] coordinates {(0.213,3500) (0.277,3500) (0.326,3500) (0.372,3500) (0.408,3500) (0.442,3500) (0.479,3521.13) (1.180,3584.23) (1.229,3676.47) (1.508,3875.97) (2.007,3952.57) (2.137,4016.06) (2.931,4098.36) (8.939,4149.38) (10.888,4201.68) (14.767,4237.29) (15.278,4237.29)};
\addplot[only marks, mark=*, mark size=1.8pt, coral, forget plot] coordinates {(15.278,4237.29)};
\addplot[const plot, evox, line width=1.0pt] coordinates {(0.000,3500) (0.052,3500) (0.076,3500) (0.348,3891.05) (0.348,3891.05) (0.515,4048.58) (0.551,4048.58) (1.116,4149.38) (1.324,4166.67) (1.324,4166.67) (4.947,4166.67)};
\addplot[only marks, mark=*, mark size=1.8pt, evox, forget plot] coordinates {(4.947,4166.67)};
\addplot[const plot, ours, line width=1.6pt] coordinates {(1.042,3500) (1.042,3500) (1.042,3500) (1.042,3500) (1.042,3500) (1.042,3597.12) (2.309,3610.11) (2.309,3623.19) (2.309,3745.32) (2.309,4000) (5.478,4032.26) (5.914,4098.36) (6.211,4115.23) (6.388,4182.23) (6.454,4189.38) (9.570,4271.68) (9.840,4274.41) (15,4274.41)};
\addplot[only marks, mark=*, mark size=1.8pt, ours, forget plot] coordinates {(15,4274.41)};
\nextgroupplot[title={LLM\_SQL}, ymin=0.66, ymax=0.735, xmax=15.0]
\addplot[const plot, arbor, line width=1.0pt, densely dashed] coordinates {(0.089,0.671994) (0.810,0.704134) (1.449,0.708514) (4.365,0.719333) (6.968,0.720538) (6.968,0.720538)};
\addplot[only marks, mark=*, mark size=1.8pt, arbor, forget plot] coordinates {(6.968,0.720538)};
\addplot[const plot, coral, line width=1.0pt] coordinates {(0.203,0.698424) (0.299,0.698913) (1.198,0.711819) (2.143,0.712634) (2.252,0.712818) (2.392,0.713675) (2.820,0.714174) (3.217,0.715564) (3.428,0.716068) (4.441,0.71631) (4.622,0.71634) (14.440,0.71634)};
\addplot[only marks, mark=*, mark size=1.8pt, coral, forget plot] coordinates {(14.440,0.71634)};
\addplot[const plot, evox, line width=1.0pt] coordinates {(0.000,0.673736) (0.114,0.709147) (0.160,0.710145) (0.540,0.710547) (0.596,0.71077) (0.596,0.7108) (0.656,0.712161) (0.731,0.714917) (1.003,0.715752) (1.003,0.7158) (1.079,0.717445) (1.329,0.717787) (1.329,0.7178) (1.785,0.718615) (4.403,0.719352) (4.661,0.7194) (7.167,0.7196)};
\addplot[only marks, mark=*, mark size=1.8pt, evox, forget plot] coordinates {(7.167,0.7196)};
\addplot[const plot, ours, line width=1.6pt] coordinates {(0.853,0.674074) (1.098,0.691811) (1.957,0.699737) (4.894,0.702841) (14.003,0.704824) (14.420,0.725125) (15,0.725125)};
\addplot[only marks, mark=*, mark size=1.8pt, ours, forget plot] coordinates {(15.0,0.725125)};
\end{groupplot}
\node[anchor=north] at ([yshift=-1.0cm]group c3r1.south) {\pgfplotslegendfromname{costlegend}};
\end{tikzpicture}%
    }
    \caption{Best score against model spend for each method on the five ADRS tasks.}
    \label{fig:cost-vs-trajectory-adrs}
\end{figure}
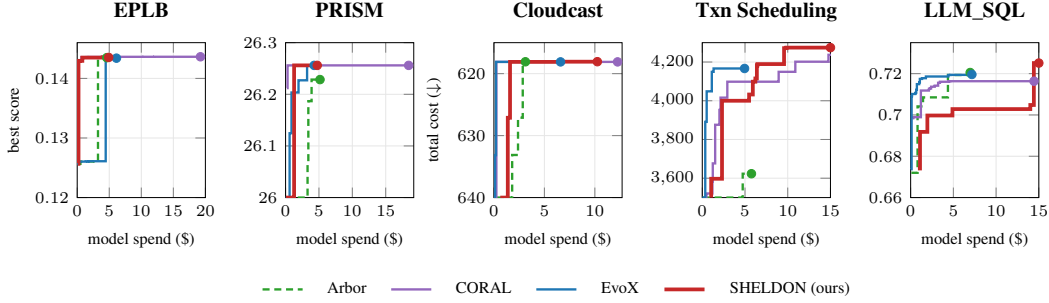

\textbf{\system{} exhibits distinct cost-performance trajectories across benchmarks}, reaching competitive solutions with little model spend on simpler tasks while continuing to improve when additional exploration remains valuable, as shown in Figure~\ref{fig:cost-vs-trajectory-adrs}.
On EPLB and PRISM, it reaches near-final performance within roughly \$1--\$1.3, whereas on transaction scheduling it continues making substantial gains through roughly \$10, ultimately reaching 4274.41. The contrast is strongest on LLM\_SQL: performance improves only gradually through about \$14 of spend before a late jump from 0.7048 to 0.7245, consistent with our observation that newly explored mechanisms can produce large improvements after an extended plateau.
Overall, the trajectories suggest that for \system{}, additional spend is not uniformly required, but can support continued exploration that yields higher-quality solutions on tasks with greater remaining headroom.

\textbf{Coupling exploration with holistic hypothesis investigation.}
\system{} pairs hypothesis exploration with holistic candidate and comparative analysis, so that relevant alternatives are not merely proposed but also investigated.
On ORCA-Bench across 30 tasks, \system{} generates hypotheses that cover 59 of 77 ground-truth causes (76.6\%), compared to 36 (46.8\%) for GoS. 
This broader coverage is accompanied by more comprehensive investigations.
\system{} tests 58 of its 59 generated hypotheses that identified the correct cause (98.3\%), whereas only 23 of the 36 causes identified by GoS (63.9\%) were investigated.
The difference therefore extends beyond proposing plausible hypotheses to rigorously investigating them.
\Explore diversifies the working hypothesis set, while the candidate and comparative analyses examine them against observed evidence.

\textbf{Exploration vs. Exploitation.}
The ground truth causes found by \system{} on ORCA-Bench arise through both exploration and exploitation.
Of the 59 ground-truth causes found, 40 are 
identified through \textsc{Explore}, while the remaining 19 emerge
through refining hypotheses already under
consideration post testing.
Thus, exploration accounts for most discoveries, but exploitation
also contributes substantially: relevant explanations can emerge
not only by opening new directions, but also by developing existing
hypotheses as additional evidence exposes their limitations.

A similar pattern appears in the optimization tasks in
Figure~\ref{fig:winning_idea_time}, where the timing of the winning ideas varies substantially across tasks. In several cases, such as PRISM and transaction scheduling, the winning region is opened early, but the winning hypothesis appears only later, \textbf{suggesting that continued exploitation within an already-discovered region is important for realizing its potential}. In contrast, for LLM SQL and Cloudcast, the winning region and winning hypothesis emerge at nearly the same time, \textbf{indicating that further exploration is what uncovers the decisive direction.} QNN provides another exploitation-heavy case: the relevant region is available from the first round, while the best score is reached only after several rounds of refinement.

\begin{figure}[t]
  \centering
  \begin{minipage}[b]{0.52\textwidth}
    \centering
    \begin{tikzpicture}
    \begin{axis}[width=3.8cm, height=3.5cm, legend cell align=left, xmin=0, xmax=9, xlabel={round}, xlabel style={font=\small}, xticklabel style={font=\footnotesize, text=muted}, xmajorgrids, major grid style={gridline, line width=0.4pt}, axis line style={draw=black!65, line width=0.5pt}, tick style={draw=none}, ytick=data, yticklabel style={font=\small}, y dir=reverse, enlarge y limits=0.06, symbolic y coords={llm_sql,txn_scheduling,cloudcast,prism,eplb,qnn}, yticklabels={{llm\_sql},{txn\_sched},{cloudcast},{prism},{eplb},{qnn}}, legend style={draw=none, fill=none, font=\scriptsize, at={(1.03,0.5)}, anchor=west, legend columns=1, row sep=-2pt}]
    \draw[track, line width=5pt, line cap=round] (axis cs:1,llm_sql) -- (axis cs:7,llm_sql);
    \draw[track, line width=5pt, line cap=round] (axis cs:1,txn_scheduling) -- (axis cs:8,txn_scheduling);
    \draw[track, line width=5pt, line cap=round] (axis cs:1,cloudcast) -- (axis cs:4,cloudcast);
    \draw[track, line width=5pt, line cap=round] (axis cs:1,prism) -- (axis cs:8,prism);
    \draw[track, line width=5pt, line cap=round] (axis cs:1,eplb) -- (axis cs:7,eplb);
    \draw[track, line width=5pt, line cap=round] (axis cs:1,qnn) -- (axis cs:8,qnn);
    \addplot[only marks, mark=square*, mark size=2.2pt, color=c3, draw=white, every mark/.append style={yshift=5pt}, forget plot] coordinates {(5,llm_sql) (1,txn_scheduling) (3,cloudcast) (1,prism) (1,eplb) (1,qnn)};
    \addlegendimage{only marks, mark=square*, mark size=2.2pt, c3}\addlegendentry{region opened}
    \addplot[only marks, mark=diamond, mark size=4.4pt, color=c2, line width=1.1pt] coordinates {(5,llm_sql) (7,txn_scheduling) (3,cloudcast) (2,prism) (7,eplb) (3,qnn)};
    \addlegendentry{best score}
    \addplot[only marks, mark=*, mark size=2pt, color=c1] coordinates {(5,llm_sql) (7,txn_scheduling) (3,cloudcast) (2,prism) (6,eplb) (1,qnn)};
    \addlegendentry{hypothesis born}
    \addlegendimage{line width=5pt, track}\addlegendentry{rounds run}
    \end{axis}
    \end{tikzpicture}
    \caption{Round in which the winning region was opened and the winning hypothesis proposed, per task.}
    \label{fig:winning_idea_time}
  \end{minipage}\hfill
  \begin{minipage}[b]{0.46\textwidth}
    \centering
    \begin{tikzpicture}
    \begin{axis}[
        width=3.4cm, height=2.2cm, scale only axis,
        xbar, bar width=8pt,
        xmin=0, xmax=1.15,
        xtick={0,0.25,0.5,0.75,1},
        xticklabels={0,25\%,50\%,75\%,100\%},
        xticklabel style={font=\footnotesize, text=inkmuted},
        xlabel={Improvement relative to seed},
        xlabel style={font=\small},
        xmajorgrids, major grid style={gridline, line width=0.4pt},
        axis line style={draw=none}, tick style={draw=none},
        symbolic y coords={cloudcast,qnn,llmsql,txn,eplb,prism},
        ytick=data,
        yticklabels={cloudcast, qnn, llm\_sql, txn\_sched, eplb, prism},
        yticklabel style={font=\small},
        enlarge y limits=0.09,
        nodes near coords={\pgfmathprintnumber[fixed,precision=0]{\pgfplotspointmeta}\%},
        nodes near coords style={font=\scriptsize, text=black, anchor=west, xshift=2pt,
                                  fill=white, inner sep=1.5pt},
        clip=false,
        point meta={x*100},
    ]
    \addplot[fill=seedless, draw=white, line width=1pt,
             postaction={pattern={Lines[angle=45,distance=4pt,line width=0.6pt]},
                         pattern color=white}]
      coordinates {(0.976,cloudcast) (0.737,qnn) (0.793,llmsql)
                   (0.955,txn) (0.966,eplb) (0.999,prism)};
    \draw[black!65, line width=0.6pt] (rel axis cs:0,0) rectangle (rel axis cs:1,1);
    \draw[withseed, dashed, line width=1.2pt]
      ({axis cs:1,cloudcast}|-{rel axis cs:0,0}) -- ({axis cs:1,prism}|-{rel axis cs:0,1})
      node[above, font=\scriptsize, text=withseed] {with seed};
    \end{axis}
    \end{tikzpicture}
    \caption{\system{} without a seed program, relative to the seeded run.}
    \label{fig:seedseedless}
  \end{minipage}
\end{figure}

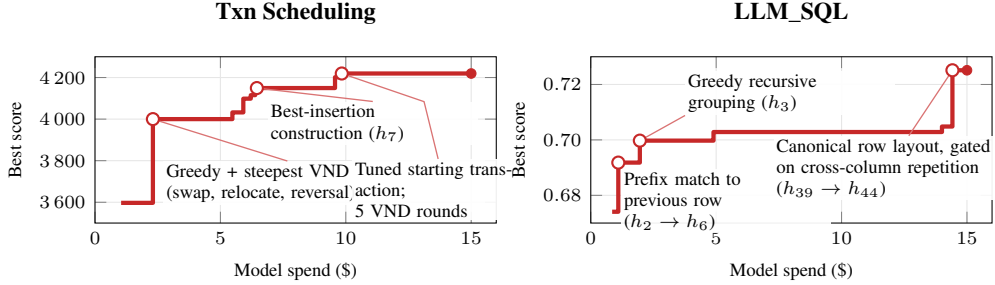
\begin{figure*}[!ht]
  \centering
  \begin{tikzpicture}[
    milestone/.style={circle, draw=ours, fill=white,
      line width=0.8pt, inner sep=1.6pt},
    annotation/.style={font=\scriptsize, align=left,
      inner sep=1pt, fill=white, fill opacity=0.94, text opacity=1},
    leader/.style={draw=ours!65, line width=0.5pt}
  ]
  \begin{groupplot}[
    group style={group size=2 by 1, horizontal sep=1.25cm},
    width=0.38\textwidth, height=2.2cm, scale only axis,
    clip=false,
    xlabel={Model spend (\$)}, ylabel={Best score}, xmin=0,
    xmajorgrids, ymajorgrids, grid style={gray!20},
    tick align=inside,
    tick label style={font=\scriptsize},
    label style={font=\scriptsize},
    title style={font=\bfseries\small, yshift=3pt},
    scaled y ticks=false,
    every axis plot/.append style={line join=round}
  ]

  \nextgroupplot[
    title={Txn Scheduling}, ymin=3500, ymax=4300, xmax=16,
    xtick={0,5,10,15}, ytick={3600,3800,4000,4200},
    yticklabel style={/pgf/number format/fixed,
      /pgf/number format/precision=0,
      /pgf/number format/1000 sep={\,}}
  ]
  \addplot[const plot, draw=ours, line width=1.6pt] coordinates {
    (1.042,3597.12) (2.309,4000) (5.478,4032.26)
    (5.914,4098.36) (6.211,4115.23) (6.388,4132.23)
    (6.454,4149.38) (9.570,4201.68) (9.840,4219.41)
    (15,4219.41)
  };
  \addplot[only marks, mark=*, mark size=1.8pt,
    draw=ours, fill=ours] coordinates {(15,4219.41)};

  \node[annotation, anchor=west, text width=3.2cm] (txn-a)
    at (axis description cs:0.17,0.24)
    {Greedy + steepest VND\\(swap, relocate, reversal)};
  \node[annotation, anchor=west, text width=2.7cm] (txn-b)
    at (axis description cs:0.43,0.60)
    {Best-insertion\\construction ($h_7$)};
  \node[annotation, anchor=west, text width=2.2cm] (txn-c)
    at (axis description cs:0.64,0.20)
    {Tuned starting transaction;\\5 VND rounds};
  \draw[leader] (axis cs:2.309,4000) -- (txn-a.north);
  \draw[leader] (axis cs:6.454,4149.38) -- (txn-b.north);
  \draw[leader] (axis cs:9.840,4219.41)
    -- (axis description cs:0.82,0.73) -- (txn-c.north);
  \node[milestone] at (axis cs:2.309,4000) {};
  \node[milestone] at (axis cs:6.454,4149.38) {};
  \node[milestone] at (axis cs:9.840,4219.41) {};

  \nextgroupplot[
    title={LLM\_SQL}, ymin=0.67, ymax=0.73, xmax=16,
    xtick={0,5,10,15}, ytick={0.68,0.70,0.72},
    yticklabel style={/pgf/number format/fixed,
      /pgf/number format/fixed zerofill,
      /pgf/number format/precision=2}
  ]
  \addplot[const plot, draw=ours, line width=1.6pt] coordinates {
    (0.853,0.674074) (1.098,0.691811) (1.957,0.699737)
    (4.894,0.702841) (14.003,0.704824) (14.420,0.725125)
    (15,0.725125)
  };
  \addplot[only marks, mark=*, mark size=1.8pt,
    draw=ours, fill=ours] coordinates {(15,0.725125)};

  \node[annotation, anchor=west, text width=2.7cm] (sql-a)
    at (axis description cs:0.075,0.13)
    {Prefix match to\\previous row\\($h_2 \rightarrow h_6$)};
  \node[annotation, anchor=west, text width=2.7cm] (sql-b)
    at (axis description cs:0.235,0.78)
    {Greedy recursive\\grouping ($h_3$)};
  \node[annotation, anchor=west, text width=3.5cm] (sql-c)
    at (axis description cs:0.455,0.33)
    {Canonical row layout, gated\\on cross-column repetition\\($h_{39} \rightarrow h_{44}$)};
  \draw[leader] (axis cs:1.098,0.691811) -- (sql-a.north west);
  \draw[leader] (axis cs:1.957,0.699737) -- (sql-b.south west);
  \draw[leader] (axis cs:14.420,0.725125) -- (sql-c.north);
  \node[milestone] at (axis cs:1.098,0.691811) {};
  \node[milestone] at (axis cs:1.957,0.699737) {};
  \node[milestone] at (axis cs:14.420,0.725125) {};
  \end{groupplot}
  \end{tikzpicture}
  \caption{Search trajectories of \system{} on transaction scheduling and
  LLM\_SQL: best score versus cumulative spend. Open circles mark annotated
  improvements, filled circles the final scores. The largest LLM\_SQL gain follows
  the gated combination of canonical row ordering with conditional grouping ($h_{39}\rightarrow h_{44}$).}
  \label{fig:sheldon-search-progress}
\end{figure*}

\begin{figure}[!t]
    \centering
    \includegraphics[width=\linewidth]{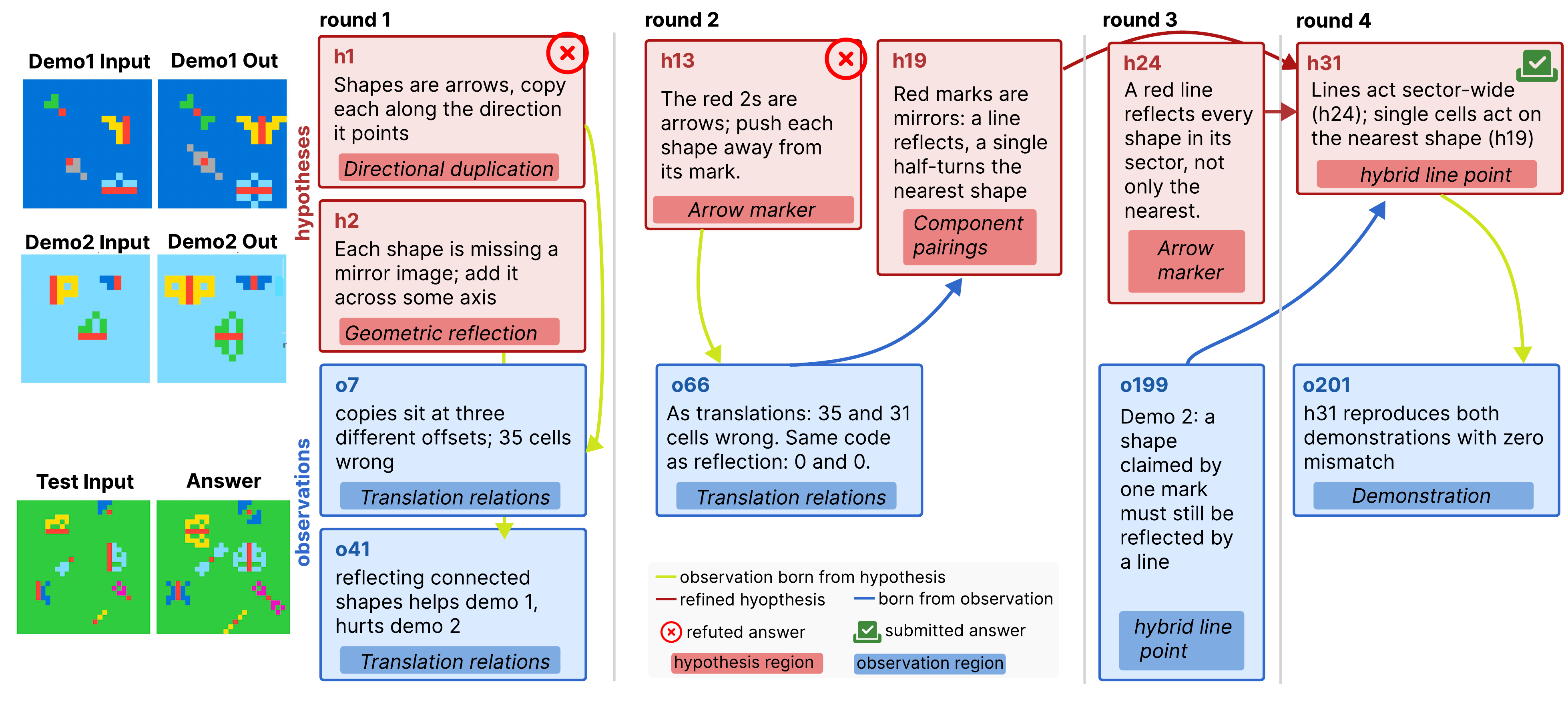}
    \caption{\system{} on an ARC-AGI-2 task. $h_1$ and $h_2$ are refuted, $h_{19}$ is generated from an observation, and $h_{31}$ combines earlier hypotheses and their observations into the submitted answer.}
\label{fig:arc-example}
\end{figure}

\textbf{How strategies evolve.}
On transaction scheduling, \system{} first explored several greedy constructions (\(h_1\)--\(h_4\)), then improved them with variable-neighborhood descent (\(h_5\)), reaching 4000, as shown in Figure~\ref{fig:sheldon-search-progress}. When this line plateaued, \textsc{Explore} introduced a new NEH-style best-insertion family (\(h_7\)), which was combined with the earlier local search in \(h_8\)--\(h_{11}\) to reach 4274. Follow-up tests showed that reversal was unnecessary (\(h_{12}\)) while the starting transaction mattered (\(h_{13}\)), narrowing the final strategy to the components that actually improved performance.
On LLM\_SQL, early progress came from prefix matching (\(h_2\), refined to \(h_6\)) and greedy recursive grouping (\(h_3\)), with further refinement to a bounded conditional grouping tree (\(h_{15}\)) reaching 0.703, as shown in Figure~\ref{fig:intro}. Although performance then plateaued, \textsc{Explore} continued opening new regions and eventually discovered a multiset-canonical layout strategy (\(h_{39}\)). Combining this new mechanism with the previously refined grouping strategy in \(h_{44}\) produced the largest improvement to 0.725.

\textbf{Minimal reliance on seed programs.} While seed programs are often provided to jumpstart optimizations, 
\textbf{\system{} is not heavily dependent on them.} As shown in Figure~\ref{fig:seedseedless}, it recovers 99.9\% of the seeded improvement on PRISM, 97.6\% on Cloudcast, 96.6\% on EPLB, and 95.5\% on transaction scheduling, approaching the gains obtained with an initial implementation. Recovery remains substantial on LLM\_SQL and QNN Circuit Topology, at 79.3\% and 73.7\%, respectively. These results demonstrate that \system{} can generate and refine effective solutions without a supplied starting program, while the larger gaps on LLM\_SQL and QNN indicate that informative initialization remains valuable on some tasks.

\textbf{A worked example on ARC-AGI-2.} Figure~\ref{fig:arc-example} traces how \system{} works on one of the tasks from ARC-AGI2 end to end.
From round 1, \system{} $h_1$ is refuted by candidate analysis, whereas $h_2$ is refined. 
In $o_{41}$, \system{} writes that ``reflecting connected shapes helps demo 1 but hurts demo2''. 
In round 3, $h_{19}$ is proposed from an observation $o_{66}$ rather than from an existing hypothesis.
Finally after several refinements and explorations, \system{} comes up with $h_{31}$ which combines two of the earlier hypotheses $h_{19}$ and $h_{24}$.
Note how $o_{201}$, identifies that the $h_{31}$ reproduces the provided demonstrations correctly and submits the artifact attached to it for submission.

The remainder of this section asks whether \system{}'s pattern of exploring different spaces holds in aggregate.
We check how broadly \system{} explores compared to the baselines, then what the observation space contributes to that breadth.
We perform analysis on tasks where a scoring function is available.

\subsection{Breadth of exploration}
\label{sec:breadth}

We follow the coverage analysis of AIM~\citep{choi2026aim} and extend it in two ways. 
AIM asks how much of the solution space a method visited: it embeds every intermediate solution the method produced, projects the embeddings to two principal components, and reports the area of their convex hull, with the mean pairwise cosine distance of the pool.
We compute both for every task, method, and run for which we hold intermediate solutions, and add a sample-size-matched hull. 
We consider a solution as any complete candidate method submitted to the task's evaluator during a run, whether or not it became the final answer. Appendix~\ref{appendix:coverage} gives the full procedure.

\begin{figure}[tp]
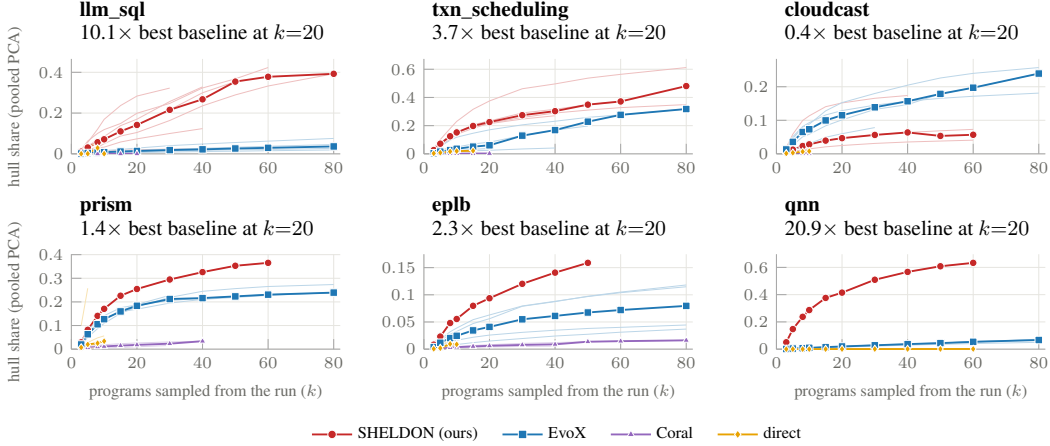

    \centering
\resizebox{\linewidth}{!}{%
%
}
    \caption{Hull share of $k$ randomly drawn programs (300 draws per $k$) in the shared PCA frame of each task; thin lines are runs, bold lines medians. Panel titles give \system{}'s ratio to the strongest baseline at $k{=}20$. A higher hull share means more novel programs generated.}
\label{fig:matched-coverage}
\end{figure}

\textbf{\system{} covers a larger region of program space.} We draw $k$ of each run's distinct solutions uniformly without replacement, compute the hull of the draw, and average over draws (300 per $k$, fixed seed), reporting the area as a share of the hull of all distinct solutions on that task. At $k{=}20$, \system{}'s median hull share exceeds the strongest baseline on five of six tasks, by factors from 1.4 to 20.9 (Figure~\ref{fig:matched-coverage}). The exception is Cloudcast, where EvoX covers more, yet \system{} gets marginally better score of 618.07 vs 618.08. This shows that, at matched cost, \system{} explore more diverse solution space. 

\pgfplotsset{compat=1.18}
\usepgfplotslibrary{groupplots,statistics}
\usetikzlibrary{plotmarks,calc}
\definecolor{revision}{HTML}{8A6FD1}
\definecolor{neutral}{HTML}{B9B8B2}
\definecolor{muted}{HTML}{6B6B66}
\definecolor{gridc}{HTML}{E6E5E0}
\definecolor{adrsband}{HTML}{F1F0EC}
\pgfplotsset{
  sheldonfig/.style={
    font=\footnotesize,
    axis lines*=left, axis line style={black!35},
    tick style={black!35}, tick label style={font=\scriptsize, text=muted},
    label style={font=\scriptsize, text=muted},
    title style={font=\footnotesize\bfseries, align=left, anchor=south west, at={(0,1)}, yshift=-2pt},
    grid=major, major grid style={gridc, line width=0.4pt},
    legend style={draw=none, fill=none, font=\scriptsize, legend columns=-1,
                  /tikz/every even column/.append style={column sep=8pt}},
    legend cell align=left,
  },
}

\begin{figure}[!t]
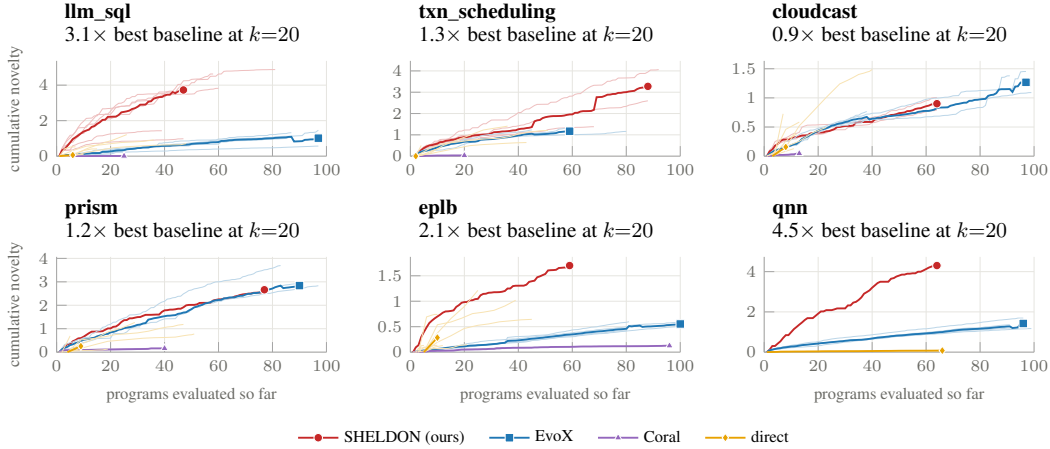

    \centering
\resizebox{\linewidth}{!}{%
%
%
}
\caption{Cumulative novelty against programs evaluated; thin lines are runs, bold lines medians, and the marker is each system's final value. Panel titles give \system{}'s ratio to the strongest baseline after 20 programs.}
\label{fig:cum-novelty}
\end{figure}

\textbf{The programs generated by \system{} keep being new.} Figure~\ref{fig:cum-novelty} tracks cumulative novelty, the running sum over evaluated programs of each program's embedding distance to the nearest program evaluated before it. A system that keeps proposing variants of what it already has accumulates novelty slowly. After 20 programs \system{} leads the strongest baseline on every task except Cloudcast, by up to $4.5\times$ on QNN.

\begin{figure}[tp]
    \centering
\resizebox{\linewidth}{!}{%
\begin{tikzpicture}
\begin{groupplot}[sheldonfig,
  group style={group size=4 by 1, horizontal sep=0.35cm, yticklabels at=edge left},
  width=4.55cm, height=6.4cm, clip=false, ymin=-0.6, ymax=11.6,
  ytick={0,...,11}, yticklabels={PPARG,PPARA,PGR,F2,CDK2,ADRB1,qnn,eplb,prism,cloudcast,txn\_scheduling,llm\_sql},
  yticklabel style={font=\scriptsize, text=black}, ymajorgrids=true,
  legend style={at={(0.5,-0.17)}, anchor=north, legend columns=1}]
\nextgroupplot[title={(a) Regions explored}, xmin=0, xmax=64, xlabel={number of regions}]
\fill[black, fill opacity=0.045] (axis cs:0,5.5) rectangle (axis cs:64,11.5);
\draw[gridc, line width=2pt] (axis cs:17.89,11) -- (axis cs:20,11);
\node[anchor=west, font=\tiny, text=muted] at (axis cs:21.50,11) {57 hyps};
\draw[gridc, line width=2pt] (axis cs:4.39,10) -- (axis cs:5,10);
\node[anchor=west, font=\tiny, text=muted] at (axis cs:6.50,10) {64 hyps};
\draw[gridc, line width=2pt] (axis cs:12.18,9) -- (axis cs:14,9);
\node[anchor=west, font=\tiny, text=muted] at (axis cs:15.50,9) {22 hyps};
\draw[gridc, line width=2pt] (axis cs:5.94,8) -- (axis cs:6,8);
\node[anchor=west, font=\tiny, text=muted] at (axis cs:7.50,8) {69 hyps};
\draw[gridc, line width=2pt] (axis cs:7.95,7) -- (axis cs:8,7);
\node[anchor=west, font=\tiny, text=muted] at (axis cs:9.50,7) {60 hyps};
\draw[gridc, line width=2pt] (axis cs:5.97,6) -- (axis cs:6,6);
\node[anchor=west, font=\tiny, text=muted] at (axis cs:7.50,6) {65 hyps};
\draw[gridc, line width=2pt] (axis cs:28.43,5) -- (axis cs:43,5);
\node[anchor=west, font=\tiny, text=muted] at (axis cs:44.50,5) {95 hyps};
\draw[gridc, line width=2pt] (axis cs:27.45,4) -- (axis cs:32,4);
\node[anchor=west, font=\tiny, text=muted] at (axis cs:33.50,4) {98 hyps};
\draw[gridc, line width=2pt] (axis cs:27.51,3) -- (axis cs:36,3);
\node[anchor=west, font=\tiny, text=muted] at (axis cs:37.50,3) {95 hyps};
\draw[gridc, line width=2pt] (axis cs:43.84,2) -- (axis cs:51,2);
\node[anchor=west, font=\tiny, text=muted] at (axis cs:52.50,2) {79 hyps};
\draw[gridc, line width=2pt] (axis cs:20.38,1) -- (axis cs:30,1);
\node[anchor=west, font=\tiny, text=muted] at (axis cs:31.50,1) {204 hyps};
\draw[gridc, line width=2pt] (axis cs:28.41,0) -- (axis cs:34,0);
\node[anchor=west, font=\tiny, text=muted] at (axis cs:35.50,0) {85 hyps};
\addplot[only marks, mark=*, mark size=2pt, neutral, mark options={draw=white}] coordinates {(20,11) (5,10) (14,9) (6,8) (8,7) (6,6) (43,5) (32,4) (36,3) (51,2) (30,1) (34,0)};
\addlegendentry{regions discovered}
\addplot[only marks, mark=*, mark size=2pt, sheldon, mark options={draw=white}] coordinates {(17.89,11) (4.391,10) (12.18,9) (5.941,8) (7.947,7) (5.97,6) (28.43,5) (27.45,4) (27.51,3) (43.84,2) (20.38,1) (28.41,0)};
\addlegendentry{regions with at least one hypothesis}
\nextgroupplot[title={(b) New-best sources}, xmin=0, xmax=14.5, xlabel={count}]
\fill[black, fill opacity=0.045] (axis cs:0,5.5) rectangle (axis cs:14.5,11.5);
\draw[gridc, line width=2pt] (axis cs:4,11) -- (axis cs:6,11);
\draw[gridc, line width=2pt] (axis cs:4,10) -- (axis cs:13,10);
\draw[gridc, line width=2pt] (axis cs:3,9) -- (axis cs:4,9);
\draw[gridc, line width=2pt] (axis cs:3,8) -- (axis cs:5,8);
\draw[gridc, line width=2pt] (axis cs:6,7) -- (axis cs:7,7);
\draw[gridc, line width=2pt] (axis cs:3,6) -- (axis cs:4,6);
\draw[gridc, line width=2pt] (axis cs:5,5) -- (axis cs:7,5);
\draw[gridc, line width=2pt] (axis cs:7,4) -- (axis cs:10,4);
\draw[gridc, line width=2pt] (axis cs:6,3) -- (axis cs:7,3);
\draw[gridc, line width=2pt] (axis cs:7,2) -- (axis cs:8,2);
\draw[gridc, line width=2pt] (axis cs:4,1) -- (axis cs:6,1);
\draw[gridc, line width=2pt] (axis cs:7,0) -- (axis cs:9,0);
\addplot[only marks, mark=*, mark size=2pt, neutral, mark options={draw=white}] coordinates {(6,11) (13,10) (4,9) (5,8) (7,7) (4,6) (7,5) (10,4) (7,3) (8,2) (6,1) (9,0)};
\addlegendentry{new run bests}
\addplot[only marks, mark=*, mark size=2pt, sheldon, mark options={draw=white}] coordinates {(4,11) (4,10) (3,9) (3,8) (6,7) (3,6) (5,5) (7,4) (6,3) (7,2) (4,1) (7,0)};
\addlegendentry{distinct source regions}
\nextgroupplot[title={(c) New-best origin}, xmin=0, xmax=1.13, xtick={0,0.25,0.5,0.75,1}, xticklabels={0,25\%,50\%,75\%,100\%}, ymajorgrids=false, xbar stacked, enlarge y limits=false]
\fill[black, fill opacity=0.045] (axis cs:0,5.5) rectangle (axis cs:1.13,11.5);
\addplot[xbar stacked, bar width=7pt, fill=sheldon, draw=white, line width=0.6pt] coordinates {(0.5000,11) (0.5385,10) (0.7500,9) (0.8000,8) (0.0000,7) (0.7500,6) (0.2857,5) (0.4000,4) (0.5714,3) (0.6250,2) (1.0000,1) (0.6667,0)};
\addlegendentry{explore}
\addplot[xbar stacked, bar width=7pt, fill=revision, draw=white, line width=0.6pt] coordinates {(0.5000,11) (0.3077,10) (0.2500,9) (0.2000,8) (1.0000,7) (0.2500,6) (0.5714,5) (0.6000,4) (0.2857,3) (0.3750,2) (0.0000,1) (0.0000,0)};
\addlegendentry{in-loop revision}
\addplot[xbar stacked, bar width=7pt, fill=neutral, draw=white, line width=0.6pt] coordinates {(0.0000,11) (0.1538,10) (0.0000,9) (0.0000,8) (0.0000,7) (0.0000,6) (0.1429,5) (0.0000,4) (0.1429,3) (0.0000,2) (0.0000,1) (0.3333,0)};
\addlegendentry{revise/refine}
\node[font=\small] at (axis cs:1.065,11) {$\star$};
\node[font=\small] at (axis cs:1.065,9) {$\star$};
\node[font=\small] at (axis cs:1.065,8) {$\star$};
\node[font=\small] at (axis cs:1.065,5) {$\star$};
\node[font=\small] at (axis cs:1.065,3) {$\star$};
\node[font=\small] at (axis cs:1.065,2) {$\star$};
\node[font=\small] at (axis cs:1.065,1) {$\star$};
\node[font=\small] at (axis cs:1.065,0) {$\star$};
\nextgroupplot[title={(d) Hypothesis depth}, xmin=0, xmax=1, xtick={0,0.25,0.5,0.75,1}, xticklabels={0,25\%,50\%,75\%,100\%}, ymajorgrids=false, xbar stacked, enlarge y limits=false]
\fill[black, fill opacity=0.045] (axis cs:0,5.5) rectangle (axis cs:1,11.5);
\addplot[xbar stacked, bar width=7pt, fill=sheldon, draw=white, line width=0.6pt] coordinates {(0.4035,11) (0.4062,10) (0.6364,9) (0.3768,8) (0.3833,7) (0.4000,6) (0.4842,5) (0.4796,4) (0.4842,3) (0.6456,2) (0.5882,1) (0.5765,0)};
\addlegendentry{depth 0 (no parent)}
\addplot[xbar stacked, bar width=7pt, fill=lightsheldon, draw=white, line width=0.6pt] coordinates {(0.4035,11) (0.3594,10) (0.2273,9) (0.2899,8) (0.4167,7) (0.3231,6) (0.3158,5) (0.2959,4) (0.3158,3) (0.3038,2) (0.2598,1) (0.3059,0)};
\addlegendentry{depth 1}
\addplot[xbar stacked, bar width=7pt, fill=neutral, draw=white, line width=0.6pt] coordinates {(0.1930,11) (0.2344,10) (0.1364,9) (0.3333,8) (0.2000,7) (0.2769,6) (0.2000,5) (0.2245,4) (0.2000,3) (0.0506,2) (0.1520,1) (0.1176,0)};
\addlegendentry{depth $\geq$2}
\end{groupplot}
\end{tikzpicture}%
}
   \caption{Origins and refinement of hypotheses. \textbf{(a)} Number of identified regions versus the number containing at least one hypothesis. \textbf{(b)} Number of hypotheses from distinct regions that improve upon the previous best. \textbf{(c)} Distribution of hypothesis origins; $\star$ marks tasks whose final best hypothesis originates from exploration. \textbf{(d)} Distribution of hypotheses by number of refinements.}
\label{fig:sheldon-structure}
\end{figure}
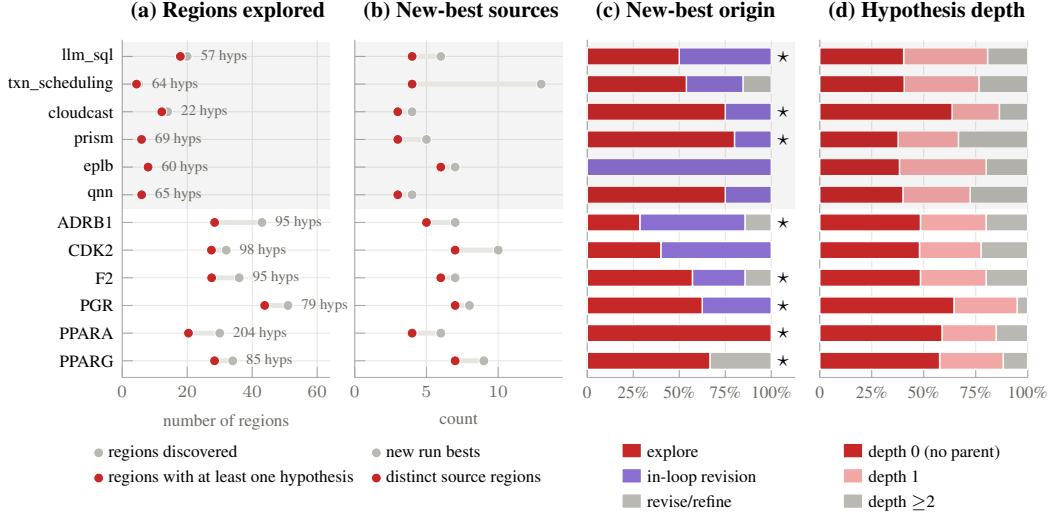

\textbf{Within \system{}, generated hypotheses span across regions and several improvements come from the newer ones.} Figure~\ref{fig:sheldon-structure} looks inside \system{}'s runs on all twelve optimization tasks, with regions as the unit. 
The number of regions that had a hypothesis is close to the effective number of regions, hence exploration is spread nearly evenly rather than concentrated in a few regions (panel a).
Each run's best score improved several times, and those improvements came from three to seven distinct regions per task (panel b).
Moreover, we classify each improvement by the origin of the hypotheses. (1) a completely new hypothesis (\Explore), (2) a local refinement, or (3) a refinement after each round (\Revise).
We observe that on 8 of 12 tasks the final best hypothesis is a new generated hypothesis, which means that exploration helps (panel c). In panel d, we show how many hypotheses are new (depth 0) and how many were refined.

\textbf{The same holds on tasks without a scoring function, where \system{} investigates several times more distinct ideas than GoS.} On DiagnosisArena and ORCA-Bench a hypothesis is a diagnosis or a root cause stated in text, so we count ideas rather than embed programs.
For each hyopthesis that went through candidate analysis with \system{}, and each candidate that GoS dispatched an expert is reduced to a short label and embedded.
We count a label as a new idea when its cosine distance to every earlier label exceeds a radius $r$ (Appendix~\ref{appendix:coverage}). 
At $r{=}0.35$, \system{} investigates more distinct ideas than GoS on 47 of 48 DiagnosisArena tasks and 26 of 27 ORCA-Bench tasks, with medians of 5 against 1.5 and 6 against 1.
The lead holds at every radius $r \in \{0.25, 0.35, 0.5\}$ (panel a shows $r{=}0.35$), and counting every hypothesis rather than only those tested gives the same direction.
GoS typically dispatches its experts to one or two frontiers: 16 of 48 DiagnosisArena tasks and 17 of 27 ORCA-Bench tasks have a single distinct label. 
The breadth of exploration with \system{} does not come at the expense of accuracy: with the same model and budget it also scores higher than GoS on both tasks. Breadth is not by itself predictive of success on a given task: across up to fourteen per-task features, neither the number of regions nor the count of distinct ideas correlates with the task's score on either benchmark, and the only feature that survives correction is the share of refuting observations on ORCA-Bench ($\rho=-0.59$), which marks the incidents the system found hard.

\pgfplotsset{compat=1.18}
\usepgfplotslibrary{groupplots}
\usetikzlibrary{plotmarks,calc}
\definecolor{orca}{HTML}{1AA874}
\definecolor{neutral}{HTML}{B9B8B2}
\definecolor{muted}{HTML}{6B6B66}
\definecolor{gridc}{HTML}{E6E5E0}
\pgfplotsset{
  sheldonfig/.style={
    font=\footnotesize,
    axis lines*=left, axis line style={black!35},
    tick style={black!35}, tick label style={font=\scriptsize, text=muted},
    label style={font=\scriptsize, text=muted},
    title style={font=\footnotesize\bfseries, align=left, anchor=south west, at={(0,1)}, yshift=-2pt},
    grid=major, major grid style={gridc, line width=0.4pt},
    legend style={draw=none, fill=none, font=\scriptsize},
    legend cell align=left,
  },
}

\begin{figure}[tp]
    \centering
\resizebox{0.85\linewidth}{!}{%
\begin{tikzpicture}
\begin{groupplot}[sheldonfig, group style={group size=3 by 1, horizontal sep=1.3cm}, width=5.0cm, height=3.5cm]
\nextgroupplot[title={(a) Distinct ideas, $r{=}0.35$}, xlabel={GoS}, ylabel={SHELDON}, xmin=0, xmax=11, ymin=0, ymax=11, xtick={0,2,...,10}, ytick={0,2,...,10}, legend style={at={(0,-0.28)}, anchor=north west, legend columns=1}]
\addplot[black!35, dashed, forget plot] coordinates {(0,0) (11,11)};
\addplot[only marks, mark=*, mark size=1.7pt, sheldon, fill opacity=0.8, mark options={draw=white, line width=0.3pt}] coordinates {(0.7928,5.175) (1.868,4.947) (1.151,6.99) (2.041,6.823) (0.7971,5.099) (1.967,3.896) (1.99,2.186) (1.83,4.888) (2.117,5.031) (0.8068,1.95) (0.9456,5.056) (2.008,3.848) (0.9653,3.84) (1.043,6.123) (3.119,5.126) (1.228,7.033) (1.892,10.21) (3.074,3.853) (1.098,5.175) (0.8964,1.834) (0.7507,3.232) (1.237,6.062) (1.899,5.053) (2.907,5.235) (2.196,7.144) (2.043,4.145) (1.986,4.777) (1.137,3.935) (1.765,2.792) (1.103,2.847) (1.937,6.857) (0.7954,6.179) (2.08,6.813) (3.216,6.898) (0.8536,4.996) (1.065,5.175) (0.8991,6.233) (2.121,5.104) (1.111,4.857) (1.859,3.022) (1.165,3.103) (1.079,4.776) (1.091,5.09) (2.16,2.934) (0.9643,6.045) (1.129,3.085) (3.189,5.085) (2.801,4.012)};\addlegendentry{DiagnosisArena}
\addplot[only marks, mark=square*, mark size=1.7pt, orca, fill opacity=0.8, mark options={draw=white, line width=0.3pt}] coordinates {(1.027,6.157) (4.849,4.16) (0.9976,6.877) (1.813,2.991) (0.9904,8.921) (1.018,2.881) (1.137,4.036) (0.9468,4.909) (0.7598,7.059) (2.014,5.041) (0.8526,2.802) (2.121,6.971) (0.9443,6.945) (1.94,6.103) (1.205,5.794) (0.9465,4.834) (0.9244,8.006) (0.924,2.952) (0.9904,6.083) (1.797,8.916) (1.023,5.849) (1.211,7.216) (2.031,6.872) (1.122,3.824) (3.224,4.89) (2.171,3.92) (1.122,4.863)};\addlegendentry{ORCA-bench}
\node[anchor=south east, font=\scriptsize, align=right] at (rel axis cs:0.97,0.03) {SHELDON more on\\47/48 DiagnosisArena\\26/27 ORCA-bench};
\nextgroupplot[title={(b) New territory}, xlabel={round}, ylabel={share}, xmin=1.7, xmax=4.3, ymin=0, ymax=0.8, xtick={2,3,4}, ytick={0,0.2,0.4,0.6,0.8}, yticklabels={0,20\%,40\%,60\%,80\%}, legend style={at={(0,-0.28)}, anchor=north west, legend columns=1}]
\addplot[sheldon, thick, mark=*, mark size=1.7pt, mark options={draw=white}] coordinates {(2,0.5737) (3,0.523) (4,0.5505)};\addlegendentry{hypotheses, DiagnosisArena}
\addplot[orca, thick, mark=square*, mark size=1.7pt, mark options={draw=white}] coordinates {(2,0.6) (3,0.7053) (4,0.7568)};\addlegendentry{hypotheses, ORCA-bench}
\addplot[sheldon, thick, densely dashed, mark=*, mark size=1.7pt, mark options={solid, fill=white}] coordinates {(2,0.1923) (3,0.1433) (4,0.1047)};\addlegendentry{observations, DiagnosisArena}
\addplot[orca, thick, densely dashed, mark=square*, mark size=1.7pt, mark options={solid, fill=white}] coordinates {(2,0.2033) (3,0.1549) (4,0.1714)};\addlegendentry{observations, ORCA-bench}
\nextgroupplot[title={(c) Reach of comparative obs.}, xlabel={distinct hypothesis regions judged}, ylabel={share of observations}, ybar, bar width=5pt, xmin=0.3, xmax=8.7, ymin=0, xtick={1,...,8}, ymax=0.42, ytick={0,0.1,0.2,0.3,0.4}, yticklabels={0,10\%,20\%,30\%,40\%}, xmajorgrids=false, legend style={at={(0.97,0.97)}, anchor=north east}]
\addplot[fill=sheldon, draw=white] coordinates {(1,0.0919) (2,0.04376) (3,0.1061) (4,0.151) (5,0.2998) (6,0.1193) (7,0.08862) (8,0.09956)};\addlegendentry{DiagnosisArena (median 5)}
\addplot[fill=orca, draw=white] coordinates {(1,0.1396) (2,0.2314) (3,0.2294) (4,0.1797) (5,0.1816) (6,0.01338) (7,0.02294) (8,0.001912)};\addlegendentry{ORCA-bench (median 3)}
\end{groupplot}
\end{tikzpicture}%
}
    \caption{Exploration on DiagnosisArena and ORCA-Bench. \textbf{(a)} Distinct ideas investigated per task by \system{} and GoS. \textbf{(b)} Share of each round's new hypotheses placed in a region opened after round 1 (solid) and of its observations filed in a region receiving its first observation that round (dashed). \textbf{(c)} Distinct hypothesis regions judged by each comparative-analysis observation.}
\label{fig:text-breadth}
\end{figure}
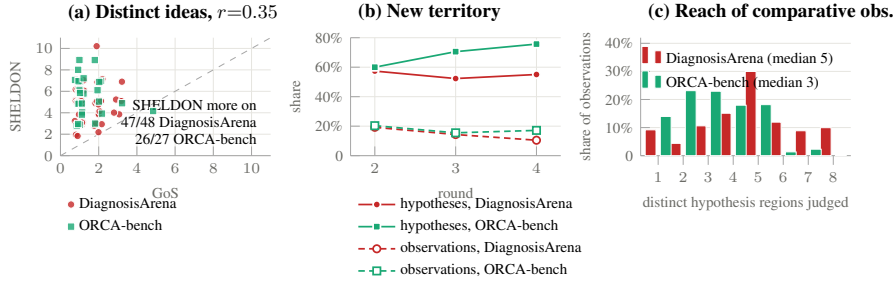

\subsection{What the observation space contributes}
\label{sec:obs-contribution}

The second analysis examines \system{}'s internal structures rather than its outputs. We track the number of task regions and the round in which each region was opened. Regions opened after round~1 represent new directions discovered by the model during exploration, rather than those specified initially.
\emph{Shared observations} are observations that contribute to the verdicts of two or more hypotheses. Such relationships cannot be captured by approaches that associate each observation with a single hypothesis, as is common in prior work such as GoS and ScientistOne.
\emph{Seeded hypotheses} are new hypotheses generated after round~1 through exploration of the observation space. These emerge when \system{}, while investigating an existing hypothesis, identifies observations that suggest new hypotheses worth pursuing.

 \textbf{Unlike prior work, our observations inform multiple hypotheses and seed new ones.} For optimization tasks, we find that 88\% of observations on ARC-AGI-2 and 82\% on QNN and ADRS evaluate two or more hypotheses (Figure~\ref{fig:observations}a). For tasks without an explicit scoring function, observations from comparative analysis constitute only 16--17\% of all observations. Despite their smaller share, these observations are disproportionately influential, accounting for 53\% and 43\% of all verdicts, respectively. Moreover, 93--98\% are cited in subsequent decisions, compared with only 9--19\% of observations from candidate analysis. We also find that a substantial fraction of observations refute at least one hypothesis (Figure~\ref{fig:observations}b), highlighting the value of actively seeking disconfirming evidence. Beyond evaluating existing hypotheses, observations also drive exploration: of the 782 ARC-AGI-2 hypotheses generated after round~1, 204 explicitly cite an observation in their rationale. In 158 cases, at least one cited observation originated from testing a different hypothesis, and in 37 cases, from analyzing a hypothesis other than the new hypothesis's parent, indicating exploration beyond the original hypothesis lineage (Figure~\ref{fig:observations}c). These figures are lower bounds, as we count only explicit citations. Consistent with these findings, the ablation in Table~\ref{tab:ablations} shows that removing comparative analysis reduces performance by 6--10 points.

\pgfplotsset{compat=1.18}
\usepgfplotslibrary{groupplots}
\usetikzlibrary{plotmarks,calc}
\definecolor{obsspace}{HTML}{1AA874}
\definecolor{revision}{HTML}{8A6FD1}
\definecolor{neutral}{HTML}{B9B8B2}
\definecolor{muted}{HTML}{6B6B66}
\definecolor{gridc}{HTML}{E6E5E0}
\pgfplotsset{
  sheldonfig/.style={
    font=\footnotesize,
    axis lines*=left, axis line style={black!35},
    tick style={black!35}, tick label style={font=\scriptsize, text=muted},
    label style={font=\scriptsize, text=muted},
    title style={font=\footnotesize\bfseries, align=left, anchor=south west, at={(0,1)}, yshift=-2pt},
    grid=major, major grid style={gridc, line width=0.4pt},
    legend style={draw=none, fill=none, font=\scriptsize},
    legend cell align=left,
  },
}

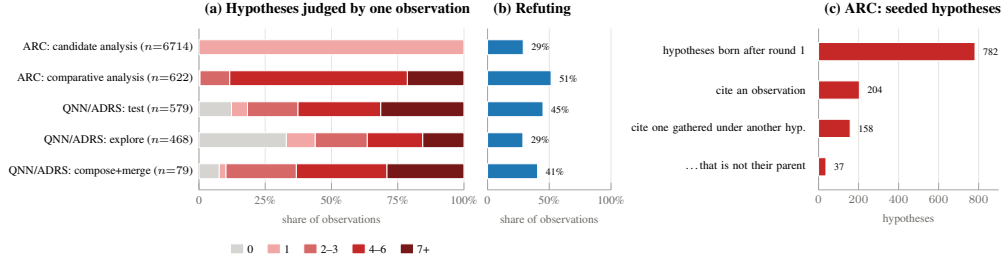
\begin{figure}[!t]
    \centering
\resizebox{0.95\linewidth}{!}{%
\begin{tikzpicture}
\begin{axis}[sheldonfig, name=a, xbar stacked, bar width=9pt, width=7.2cm, height=5.0cm,
  xmin=0, xmax=1, ymin=-0.6, ymax=4.6, ytick={0,...,4}, enlarge y limits=false,
  yticklabels={QNN/ADRS: compose+merge ($n{=}79$),QNN/ADRS: explore ($n{=}468$),QNN/ADRS: test ($n{=}579$),ARC: comparative analysis ($n{=}622$),ARC: candidate analysis ($n{=}6714$)},
  yticklabel style={font=\scriptsize, text=black}, ytick style={draw=none}, ymajorgrids=false,
  xtick={0,0.25,0.5,0.75,1}, xticklabels={0,25\%,50\%,75\%,100\%},
  xlabel={share of observations}, title={(a) Hypotheses judged by one observation},
  legend style={at={(0.5,-0.3)}, anchor=north, legend columns=-1, /tikz/every even column/.append style={column sep=6pt}}]
\addplot[fill=neutral!60, draw=white, line width=0.6pt] coordinates {(0.0000,4) (0.0000,3) (0.1226,2) (0.3291,1) (0.0759,0)};\addlegendentry{0}
\addplot[fill=lightsheldon, draw=white, line width=0.6pt] coordinates {(1.0000,4) (0.0032,3) (0.0604,2) (0.1090,1) (0.0253,0)};\addlegendentry{1}
\addplot[fill=sheldon!70, draw=white, line width=0.6pt] coordinates {(0.0000,4) (0.1125,3) (0.1900,2) (0.1966,1) (0.2658,0)};\addlegendentry{2--3}
\addplot[fill=sheldon, draw=white, line width=0.6pt] coordinates {(0.0000,4) (0.6704,3) (0.3126,2) (0.2094,1) (0.3418,0)};\addlegendentry{4--6}
\addplot[fill=sheldon!60!black, draw=white, line width=0.6pt] coordinates {(0.0000,4) (0.2138,3) (0.3143,2) (0.1560,1) (0.2911,0)};\addlegendentry{7+}
\end{axis}
\begin{axis}[sheldonfig, name=b, at={($(a.east)+(0.5cm,0)$)}, anchor=west, xbar, bar width=9pt, width=4.2cm, height=5.0cm,
  xmin=0, xmax=1, ymin=-0.6, ymax=4.6, ytick={0,...,4}, yticklabels={}, ytick style={draw=none}, ymajorgrids=false,
  xtick={0,0.5,1}, xticklabels={0,50\%,100\%}, xlabel={share of observations}, title={(b) Refuting}, clip=false]
\addplot[fill=evox, draw=white] coordinates {(0.2891,4) (0.5145,3) (0.4491,2) (0.2863,1) (0.4051,0)};
\node[anchor=west, font=\tiny] at (axis cs:0.309,4) {29\%};
\node[anchor=west, font=\tiny] at (axis cs:0.534,3) {51\%};
\node[anchor=west, font=\tiny] at (axis cs:0.469,2) {45\%};
\node[anchor=west, font=\tiny] at (axis cs:0.306,1) {29\%};
\node[anchor=west, font=\tiny] at (axis cs:0.425,0) {41\%};
\end{axis}
\begin{axis}[sheldonfig, name=c, at={($(b.east)+(4.4cm,0)$)}, anchor=west, xbar, bar width=11pt, width=5.4cm, height=5.0cm,
  xmin=0, xmax=900, ymin=-0.6, ymax=3.6, ytick={0,...,3}, ytick style={draw=none}, ymajorgrids=false, clip=false,
  yticklabels={{\ldots that is not their parent},{cite one gathered under another hyp.},{cite an observation},{hypotheses born after round 1}},
  yticklabel style={font=\scriptsize, text=black, align=right, text width=4.0cm},
  xlabel={hypotheses}, title={(c) ARC: seeded hypotheses}]
\addplot[fill=sheldon, draw=white] coordinates {(782,3) (204,2) (158,1) (37,0)};
\node[anchor=west, font=\tiny] at (axis cs:794,3) {782};
\node[anchor=west, font=\tiny] at (axis cs:216,2) {204};
\node[anchor=west, font=\tiny] at (axis cs:170,1) {158};
\node[anchor=west, font=\tiny] at (axis cs:49,0) {37};
\end{axis}
\end{tikzpicture}%
}
    \caption{Observations as shared evidence. \textbf{(a)} Hypotheses judged per observation, by the stage that filed it. \textbf{(b)} Share of observations that refute at least one hypothesis. \textbf{(c)} ARC-AGI-2 hypotheses born after round 1, by whether their rationale cites an observation and where that observation came from.}
\label{fig:observations}
\end{figure}

\begin{figure}[tp]
    \centering
    \resizebox{0.85\linewidth}{!}{%
\begin{tikzpicture}
\begin{groupplot}[sheldonfig, group style={group size=3 by 1, horizontal sep=1.35cm}, width=5.6cm, height=4.6cm]
\nextgroupplot[title={(a) Observations in new territory}, xlabel={round}, ylabel={filed in a region opened that round}, xmin=1.5, xmax=10.5, ymin=0, ymax=0.75, ytick={0,0.25,0.5,0.75}, yticklabels={0,25\%,50\%,75\%}, legend style={at={(0.97,0.97)}, anchor=north east}]
\addplot[sheldon, thick, mark=*, mark size=1.7pt, mark options={draw=white}] coordinates {(2,0.2646) (3,0.139) (4,0.168) (5,0.1498) (6,0.1703)};\addlegendentry{ARC-AGI-2}
\addplot[obsspace, thick, mark=square*, mark size=1.7pt, mark options={draw=white}] coordinates {(2,0.2712) (3,0.04255) (4,0.08889) (6,0.2353) (9,0.1667)};\addlegendentry{QNN}
\addplot[black!55, thick, mark=triangle*, mark size=1.7pt, mark options={draw=white}] coordinates {(2,0.5263) (3,0.2162) (4,0.2989) (5,0.48) (6,0.2381) (7,0.4615) (8,0.3125) (9,0.35) (10,0.3939)};\addlegendentry{ADRS}
\nextgroupplot[title={(b) Reach of comparative obs.}, xlabel={distinct hypothesis regions judged}, ylabel={observations}, ybar, bar width=8pt, xmin=0.3, xmax=9.7, ymin=0, xtick={1,...,9}, xmajorgrids=false]
\addplot[fill=sheldon, draw=white] coordinates {(1,5) (2,15) (3,62) (4,65) (5,171) (6,110) (7,56) (8,36) (9,8)};
\node[anchor=north east, font=\scriptsize, align=right] at (rel axis cs:0.98,0.97) {median 5\\96\% judge $\geq$3};
\nextgroupplot[title={(c) Empty regions, filled later}, ylabel={share of observation regions}, ybar, bar width=10pt, xmin=-0.6, xmax=2.6, ymin=0, ymax=0.3, xtick={0,1,2}, xticklabels={ARC-AGI-2,QNN,ADRS}, ytick={0,0.1,0.2,0.3}, yticklabels={0,10\%,20\%,30\%}, xmajorgrids=false, legend style={at={(0.97,0.97)}, anchor=north east}]
\addplot[fill=lightsheldon, draw=white] coordinates {(0,0.1503) (1,0.1798) (2,0.1019)};\addlegendentry{opened with no observation}
\addplot[fill=sheldon, draw=white] coordinates {(0,0.0699) (1,0.1348) (2,0.0583)};\addlegendentry{\ldots and filled in a later round}
\end{groupplot}
\end{tikzpicture}%
}

    \caption{Observations and exploration. \textbf{(a)} Share of each round's observations filed in a region opened that round. \textbf{(b)} Distinct hypothesis regions judged by each comparative-analysis observation on ARC-AGI-2. \textbf{(c)} Observation regions opened empty, and the subset filled in a later round.}
\label{fig:obs-exploration}
\end{figure}
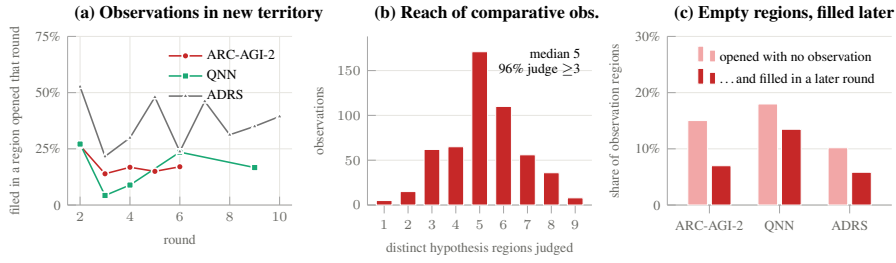

\textbf{The observation space continues to expand, connects distinct hypothesis regions, and exposes gaps for future exploration.} New observation regions emerge throughout the search: after round~1, 14--27\% of ARC-AGI-2 observations, 22--53\% of ADRS observations, and 10--20\% of DiagnosisArena and ORCA-Bench observations belong to regions opened in the same round (Figure~\ref{fig:obs-exploration}a and Figure~\ref{fig:text-breadth}b). Observations also connect hypotheses across otherwise separate regions. Each comparative-analysis observation evaluates hypotheses from a median of five distinct regions on ARC-AGI-2 and DiagnosisArena, and three on ORCA-Bench; 96\%, 86\%, and 63\% of these observations, respectively, span at least three regions (Figure~\ref{fig:obs-exploration}b and Figure~\ref{fig:text-breadth}c). Thus, a single observation can provide evidence about hypotheses that are organized separately in the hypothesis space. Finally, the observation space makes unexplored directions explicit: some regions are opened before any observations are collected, and many are populated in subsequent rounds, including 12 of 16 initially empty regions on QNN (Figure~\ref{fig:obs-exploration}c).

\FloatBarrier

\section{Conclusion}
We introduced \logosystem{}, a scientific reasoning harness that jointly searches over hypothesis and observation spaces.
By treating observations as first-class objects, \system{} enables more systematic exploration, comparison, falsification, and refinement of competing explanations.
Across diagnosis, abstract reasoning, and optimization tasks, \system{} consistently outperforms strong agentic baselines, while ablations show that explicitly modeling the observation space is critical to its performance.
These results suggest coupling hypothesis and observation search is a promising direction for building more scientific and general-purpose reasoning agents.

\section*{Acknowledgment}
We thank Dr. Lisa McPherson for insightful discussions on incorporating scientific reasoning into a harness for deep research on scientific topics.
We would also like to thank Jiuding Sun, Nikil Roashan Selvam, Prof. Yanzhe ``Sanju'' Zhang, Dr. Aryaman Arora, Shicheng Liu, Nishna Aleti, Yucheng Jiang and other members of the Stanford Open Virtual Assistant Lab (OVAL) and NLP group for their feedback and support.

This work is supported in part by the Verdant Foundation, the Hasso Plattner Institute, Itaú Unibanco, BMO Financial Group, the Stanford Human-Centered Artificial Intelligence (HAI) Institute, and the Brown Institute for Media Innovation. We acknowledge the National Artificial Intelligence Research Resource (NAIRR) Pilot and Microsoft Azure for contributing to the results in this work.

\subsection*{AI use statement}

In this work, we used generative AI tools to help with implementing
methods (coding tasks). We have not used generative AI tools to formulate
mathematical claims, design or provide feedback on research methodology
or experiments, propose or refine hypotheses, interpret results, or
support qualitative or thematic data analysis. The remaining
required-disclosure tasks, including generating synthetic datasets,
developing theoretical models or conceptual frameworks, providing
critical ingredients for mathematical proofs, assisting in the writing
of proofs, cleaning or reformatting datasets, and translation, are not
applicable to this work.

Additionally, we used generative AI tools to create or edit software
code and to polish manuscript writing for clarity and readability.
We have reviewed all AI-assisted work. LLM-generated code was verified
and tested for correctness by authors, and all AI-assisted edits to
the manuscript were also reviewed by the authors. We take responsibility for
the final content of this work, including text, claims, or artifacts
produced with the aid of generative AI.

\subsection*{Ethics statement}

Our evaluation includes medical diagnosis benchmarks, but these experiments are intended only to study the reasoning behavior of the proposed system in controlled benchmark settings. We do not make claims about \system{}'s performance in real-world clinical diagnosis, where cases may be substantially more complex and involve additional sources of uncertainty, incomplete information, workflow constraints, and patient-specific considerations. The system is not evaluated as a clinical decision-making tool, and the results should not be interpreted as evidence of suitability for medical use.

\subsection*{Reproducibility statement}

Code, data, and artifacts produced for and accompanying this paper are available at \url{https://github.com/stanford-oval/sheldon}. We provide the prompts used in Appendix~\ref{appendix:prompts}, and benchmark and baseline details in Section~\ref{sec:experiments} and Appendix~\ref{appendix:benchmarks}.

\bibliography{iclr2027_conference}
\bibliographystyle{iclr2027_conference}

\newpage
\appendix

\startcontents[appendices]

\section*{Appendix Contents}
\printcontents[appendices]{l}{1}{\setcounter{tocdepth}{2}}

\vspace{1em}

\section{Additional Experiment Setup}
\label{appendix:baselines}

\paragraph{Shared harness.} \system{}, ReAct, Codex, and GoS run inside one harness that gives every method the same task statement, the same tools, the same per-task budget, and the same model (GPT-5.6-terra, high reasoning effort). A tool is offered to a method as an ordinary function (``probe''); the set of probes is fixed per benchmark (a shell and the evaluator on ADRS and QNN; the docking oracle, RDKit descriptors, and a library-similarity search on molecular optimization; a shell in the source tree with the telemetry URLs on ORCA-Bench; the EviDx patient-record queries on DiagnosisArena; a Python sandbox on ARC-AGI-2). On the optimization benchmarks a method is scored on the best program or molecule it \emph{measured}, re-evaluated once in the benchmark's test mode, rather than on the artifact it happened to hand in last. All ablations run for eight rounds, the same as the full \system{} setting.

\paragraph{ReAct.} A single agent following the Thought / Action / Observation text protocol~\citep{yao2022react} over the same probes, with the system prompt ``You solve problems carefully and end with a clearly marked final answer. If you have tools, use them.'' and the protocol instruction below appended to the task statement. We require at least 20 iterations per run and tell the model the remaining step count; when the step ceiling is reached the protocol is withdrawn and the model is asked once more for a final answer.

\begin{lstlisting}[style=prompt]
You have these tools:
- (*@\promptvar{tool name}@*): (*@\promptvar{description}@*)
  arguments: (*@\promptvar{JSON schema}@*)

Work in this format, one step at a time:

Thought: what you are reasoning about
Action: the tool name, exactly as written above
Action Input: a JSON object of arguments

Stop after `Action Input:` -- the observation is filled in for you. When you are done, reply with `Final Answer:` and nothing after it but the answer.

Thought:
\end{lstlisting}

\paragraph{Codex.} The Codex CLI (\texttt{codex-cli} v0.128.0) run as \texttt{codex exec} in research mode with high reasoning effort on the same task statement, with the same probes served to it over MCP in addition to its own shell and file tools. Each run executes in a bubblewrap sandbox that mounts only the task workspace, so the agent cannot read evaluator sources, answer files, or earlier runs' outputs. Because an external process cannot be stopped by the harness's budget accounting, a watchdog polls Codex's session log and terminates the process at the per-task budget; the budget is also stated in the prompt so the agent can pace itself.

\paragraph{GoS.} We run the authors' released implementation~\citep{luo2026graph} with four changes that remove advantages or handicaps unrelated to its orchestration: (i) its expert sub-agents run on the same Codex engine as our Codex baseline, with the same probes; (ii) the expert output limit is raised from 4{,}096 to 16{,}000 tokens to match the other methods; (iii) the per-case step limits are raised to \texttt{session\_max\_steps}~=~10 and \texttt{max\_retrieval\_steps}~=~20 so that GoS is matched to \system{} on tool calls (its state-machine limit is set to 6 to preserve the authors' ratio; \texttt{gap\_delta}~=~0.3 and \texttt{min\_support}~=~3 are unchanged); (iv) on ORCA-Bench the role prompts are rewritten to remove benchmark-specific knowledge the originals contained (the answer class, the winning search strategy, and the fact that faults announce themselves in log text), keeping the authors' role descriptions and per-expert priority orderings. The DiagnosisArena prompts are used verbatim. GoS requires domain-specific role prompts and is therefore evaluated only on DiagnosisArena and ORCA-Bench.

\paragraph{EvoX.} The meta-evolutionary method requires a scoring function. On ARC-AGI-2 it uses the training input-output pairs of each task as its scoring function; on ADRS, QNN, and molecular optimization it uses the benchmark evaluator, under the same evaluation and dollar caps as \system{}.

\paragraph{Arbor, CORAL, ScientistOne, and RLM.} Arbor and CORAL require an initial artifact and a scoring function and are run on the optimization benchmarks only; RLM is run on the four benchmarks of Table~\ref{tab:diagnostic-reasoning-results}. ScientistOne's framework is not public, so we evaluate its released ADRS programs under our evaluator and hardware; its cost is therefore not available.

\paragraph{Evaluation without a seed program.}
For evaluation without a seed program, we instantiate \system{} in the same way as in the setting with a seed program, except that we remove the initial seed program provided to the system. We retain the surrounding engineering scaffolding, such as instructions for which tools to call and how to submit the final answer, but provide no algorithmic content.
Figure~\ref{fig:seedseedless} reports, for each task, the fraction of the seeded improvement that the seedless run recovers,
\[
\frac{\mathrm{score}(\text{without seed}) - \mathrm{score}(\text{seed program})}{\mathrm{score}(\text{with seed}) - \mathrm{score}(\text{seed program})},
\]
where $\mathrm{score}(\text{seed program})$ is the score of the provided seed program before any optimization, and $\mathrm{score}(\text{with seed})$ is the best score \system{} reaches when started from it. A value of 1 means the seedless run matches the seeded run.

\section{Benchmarks}
\label{appendix:benchmarks}
\subsection{Evaluation Hardware}
Since system optimization tasks depend on the hardware, we evaluate all baselines on two Intel Xeon E5-2660 processors at 2.20\,GHz, with eight cores and two threads per core per socket, yielding 16 physical cores, 32 logical CPUs, and two NUMA nodes.

\subsection{Setup}
\label{appendix:benchmark-setup}

\paragraph{ARC-AGI-2.} We follow the official setup, defined at \url{https://arcprize.org/arc-agi/2}, and evaluate on the public evaluation set of 120 tasks.
We use a budget of \$2.5 per task for all baselines and for \system{}.

\begin{figure}[ht]
    \centering
    \includegraphics[width=0.8\linewidth]{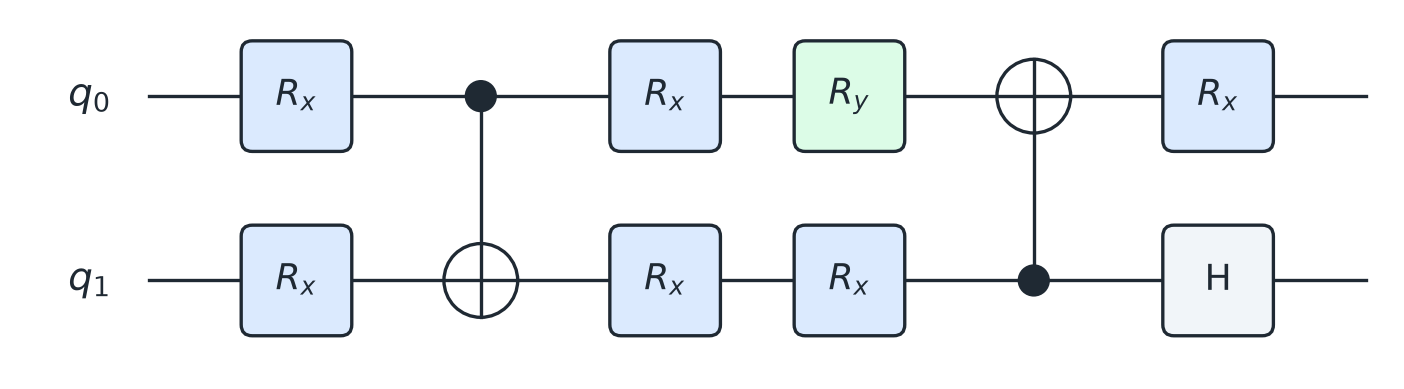}
    \caption{The two-qubit QNN circuit topology discovered by \system{} on the QNN task (held-out accuracy 1.00): ten gates at depth six, with seven trainable rotations. Each qubit first receives an $R_x$ rotation, a CNOT entangles $q_0$ and $q_1$, a second layer of rotations follows ($R_x$ on both qubits, then $R_y$ on $q_0$ and $R_x$ on $q_1$), a reverse CNOT re-entangles them, and the circuit closes with $R_x$ on $q_0$ and a Hadamard on $q_1$. Inputs are angle-encoded before the ansatz and the evaluator fits the rotation angles.}
    \label{fig:qnn-topology}
\end{figure}

\paragraph{ADRS and QNN.} We follow the official setup provided by~\citet{liu2026skydiscover} at \url{https://github.com/skydiscover-ai/skydiscover/tree/main/benchmarks}: each task supplies a working but weak seed program, a system prompt stating the problem, and an evaluator that returns a \texttt{combined\_score}. We impose a \$15 budget and allow a maximum of 100 evaluations (the upstream \texttt{max\_iterations}) for each baseline and \system{}. Programs that do not complete the whole task (a PRISM placement that fails on some configurations, an invalid transaction schedule) are rejected rather than scored. The ``Human best'' row of Table~\ref{table:adrs} is the best human-designed method for each task as reported in Table~3 of~\citet{liu2026evox}; those values were measured on the EvoX authors' hardware, and the EPLB and LLM\_SQL scores include a runtime term, so they are indicative rather than directly comparable to our rows. The metrics reported in Table~\ref{table:adrs} are the evaluators' own:
\begin{itemize}\setlength\itemsep{2pt}
\item \textbf{EPLB} (Mixture-of-Experts load balancing): $\tfrac{1}{2}(b + 0.002/t)$, where $b$ is expert-level balancedness (mean expert load divided by maximum expert load, averaged over the workload's layers) and $t$ is the mean simulated inference time in seconds; higher is better.
\item \textbf{PRISM} (LLM model placement on 80\,GB GPUs): $1/\overline{\kappa} + r$ over 50 generated placement problems, where $\kappa$ is the maximum per-GPU KV-cache pressure, $\sum_m \mathrm{req\_rate}_m/\mathrm{SLO}_m$ divided by the GPU memory left after model weights, and $r$ is the fraction of problems solved; higher is better.
\item \textbf{Cloudcast} (multi-cloud broadcast): total egress cost in USD of the broadcast topologies over the five network configurations; lower is better. (The evaluator exposes this as $1/(1+\mathrm{cost})$; we report the cost.)
\item \textbf{Transaction scheduling}: $10^6/(1+\mathrm{makespan})$ for the schedule's makespan on the provided workload; higher is better, so 4274 corresponds to a makespan of about 233.
\item \textbf{LLM\_SQL} (column reordering for prefix caching): $0.95\,h + 0.05\,(12-\min(12,t))/12$, where $h$ is the prefix-cache hit rate averaged over five datasets and $t$ the reordering runtime in seconds; higher is better.
\item \textbf{QNN circuit topology}: held-out binary classification accuracy in $[0,1]$ of a two-qubit quantum neural network whose gate sequence (RX, RY, RZ, CNOT, H; at most 24 gates) is the artifact. Inputs are angle-encoded 2D points, the evaluator fits the rotation parameters itself, and readout is $\langle Z\rangle$ on qubit 0. The seed program scores about 0.70 and the benchmark's hand-written reference topology about 0.917; \system{}'s ten-gate circuit (Figure~\ref{fig:qnn-topology}) reaches 1.00.
\end{itemize}

\begin{figure}[ht]
    \centering
    \includegraphics[width=\linewidth]{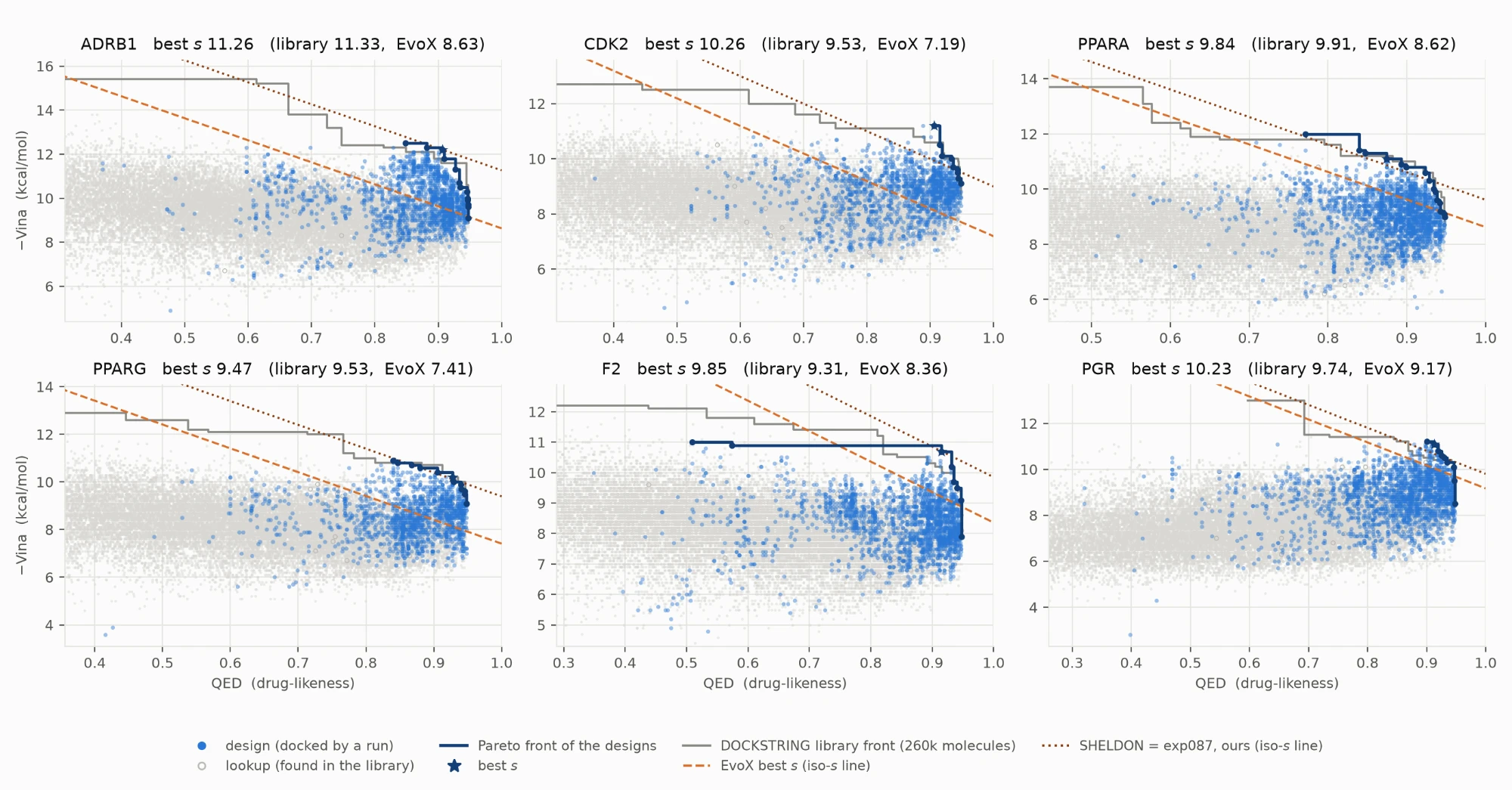}
    \caption{Molecular optimization on the six DOCKSTRING targets: $-\mathrm{Vina}$ (binding, higher is better) against QED (drug-likeness) for every molecule docked by \system{} (blue), with its Pareto front (dark blue), the best molecule by $s$ (star), and the Pareto front of the 260k-molecule DOCKSTRING library (grey). Hollow markers are molecules that were already in the library and are excluded from the fronts. Dotted and dashed lines are the level sets $s=\mathrm{const}$ through \system{}'s and EvoX's best molecules, respectively; everything above a line beats that score.}
    \label{fig:dockstring-pareto}
\end{figure}

\paragraph{Molecular optimization.} We use the dataset provided by~\citet{garcia2022dockstring} and the official PyPI package \texttt{dockstring} 0.3.4, which bundles AutoDock Vina 1.1.2 and the prepared receptors.
For the task formulation we follow~\citet{lee2025feedback}: for each of the six targets (ADRB1, PGR, PPARA, PPARG, CDK2, F2) the agent proposes molecules as SMILES strings and maximizes $s = -\mathrm{Vina} - 10\,(1-\mathrm{QED})$, where Vina is the predicted binding free energy in kcal/mol and QED is RDKit's drug-likeness in $[0,1]$. Every method receives the docking oracle, RDKit descriptors, and a similarity search over the DOCKSTRING library as tools. We permit up to 1,000 docking evaluations per target, similar to~\citet{lee2025feedback}, subject to a \$40 budget cap, and score each method on the best molecule it docked. The ``library best'' row of Table~\ref{tab:docking_results} is the highest value of $s$ among the 260,155 compounds of the DOCKSTRING database for that target (ADRB1 11.330, PGR 9.742, PPARA 9.907, PPARG 9.529, CDK2 9.534, F2 9.311), matching the ``Best Molecule'' row of Table~3 in~\citet{lee2025feedback}.
Figure~\ref{fig:dockstring-pareto} shows where the search lands for each target: every molecule docked by \system{} in the Vina--QED plane, its Pareto front, and the front of the DOCKSTRING library. Because the objective is linear in both axes, a level set of $s$ is a straight line, and the dotted line through the best molecule can be compared directly with the dashed line through EvoX's best. \system{} pushes toward the drug-like corner (QED above 0.9) and beats the library front there on CDK2, F2, and PGR.

\paragraph{DiagnosisArena.} We follow the interactive setup of EviDx~\citep{zeng2026evidx} for
DiagnosisArena~\citep{zhu2026diagnosisarena}: each case opens with the patient's
presentation, and the rest of the record is available through queries to the
environment. We report EviDx's open-ended diagnosis metric. We evaluate on a random
subset of 50 cases. We provide a budget of \$10 for each task to all baselines and \system{}.

\paragraph{Task ids.} We evaluate on the following DiagnosisArena cases (case numbers):
42, 64, 76, 94, 96, 102, 128, 144, 196, 210, 257, 266, 324, 328, 334, 367, 395, 415, 445, 454, 459, 489, 506, 517, 534, 546, 554, 561, 565, 566, 582, 617, 618, 626, 634, 656, 685, 721, 723, 724, 728, 737, 777, 819, 825, 845, 848, 865, 902, 912.

\paragraph{ORCA-Bench.} In ORCA-Bench~\citep{gong2026orca}, a task's ground truth can
include faults that began after the incident was reported. The telemetry the agent
investigates may contain them, but an operator responding at reporting time could not
have known about them, so requiring them would be unfair and leaves the task
underspecified. We therefore keep only the root causes that occurred before the
reporting time. For the same reason, the task description states that the answer
should include all plausible causes from the start of the day up to the reporting
time. We evaluate on a subset of 30 incidents, stratified by difficulty, and report
RCA accuracy as computed by ORCA-Bench's verifier: an incident counts as solved when
the report names every root cause in its ground truth. The task description, which is
the same for every method, is shown below; values filled in per task are shown as
\promptvar{name}.

\begin{lstlisting}[style=prompt]
# Root Cause Analysis

You are an expert site reliability engineer. The current time is (*@\promptvar{current time}@*).

## Task Description

The following issue was reported at (*@\promptvar{report time}@*):

(*@\promptvar{reported issue}@*)

NOTE:
* The time that an issue was reported is not necessarily the same as the time the incident actually began. Your task is to pinpoint the root cause events and the times of their occurrence.
* There may be multiple root causes or no root cause at all.
* The user may have encountered the problem at any point that day before or during the reported time.

## Source Code

The source code for the system is available at `/app/opentelemetry-demo`.

## Telemetry Tools

Please use the following tools:
- Grafana HTTP API (available at $GRAFANA_URL with username `admin` and password `admin`; the URL is set in the `GRAFANA_URL` environment variable)

## Instructions

First determine whether an incident actually occurred. If no incident occurred, please write an empty file to `/app/report.md`. If an incident occurred, please write an incident RCA report to `/app/report.md` naming the root cause events.

---

## Context

The following is not a requirement of this task; it is how incident RCA reports are conventionally written.

Summary: Focus on what DID happen and the observable impact -- mention request failures if they occurred, but do NOT mention the absence of failures or errors as this can be misread and cause confusion.

Timeline: A chronological narrative of ONLY what went wrong, reconstructed from the provided telemetry. Each entry should be a single succinct line. Cover higher-level events -- when a change was introduced, when symptoms started, when a change was reverted, when symptoms subsided -- not individual error occurrences. The first entry should be the earliest anomaly visible in the provided telemetry; this establishes the incident start time.

For each entry:
- State the time (UTC), the service(s) involved, and what happened
- Quantify where possible (error rate, duration, number of affected requests)
- Do NOT include normal/healthy behavior, baseline noise, or things that worked correctly

Example format:
```
09:44 UTC -- analytics-job deployed with no memory limit set
14:17 UTC -- redis-cache-0 OOMKilled; node memory at 97%
14:18 UTC -- inventory-svc cache misses begin; all requests fall through to Postgres
14:18 UTC -- order-svc goroutine pool exhaustion begins; 503 error rate climbs to 14%
14:29 UTC -- redis-cache-0 rescheduled; error rate returns to baseline
```

Causal chain: Each step must be grounded in the provided telemetry. Cite only the telemetry types relevant to that step, using the format below for each.

   - Metrics: one sentence describing the baseline value (or typical range), what it changed to, and at what UTC timestamp. Where relevant, include rate of change or duration. Follow with the exact PromQL query that would surface it.
     Example -- "Error rate on order-svc was 0.1% at 14:17 UTC, jumping to 14% at 14:18 UTC. PromQL: rate(http_requests_total{service="order-svc",status=~"5.."}[1m])"

   - Traces: the cascade from the highest-level operation down to the deepest failing span, followed by representative trace IDs from the provided data.
     Example -- "checkout request -> order-svc goroutine exhaustion -> inventory-svc Redis GET timeout. Trace IDs: 4bf92f3577b34da6, a3ce929d0e0e4736"

   - Logs: one sentence with service, endpoint, error, and time range, followed by the exact Lucene query that would retrieve the relevant log lines.
     Example -- "Between 14:18:02 and 14:29:11 UTC, POST /api/checkout returned 503s. Lucene: service:order-svc AND status:503 AND endpoint:\"/api/checkout\""

Never accept "human error" or "operator mistake" as a terminal answer. Ask why the human was in a position to make that error. Stop when you reach a cause that is: (a) actionable -- something a team can change, or (b) a process or system gap that allowed the failure to propagate. Only include steps that are directly supported by the provided telemetry. If the causal chain cannot be completed with the available evidence, stop and note what additional telemetry would be needed to continue. If the incident has multiple contributing causes that cannot fit a single chain, say so explicitly rather than forcing one.
\end{lstlisting}

We provide a budget of \$10 for each task to all baselines and \system{}.

These settings are the same for \system{} and
every baseline: all methods receive the same task descriptions and the same
information about each task, and are graded against the same ground truth.

\paragraph{Task ids.} We use the following ORCA-Bench task ids:
{\scriptsize
\begin{itemize}\setlength\itemsep{0pt}
\item \texttt{38b393f4b9fee541}
\item \texttt{d6-big-productcatalogfailure-on-05-easy\_ttd1440m\_day\_same-day}
\item \texttt{d1-i2-adfailure-on-03-easy\_ttd480m\_period\_same-period}
\item \texttt{d5-d4dep-cartfailure-on-03-easy\_ttd480m\_period\_same-period}
\item \texttt{d4-d3dep-productcatalogfailure-on-11-easy\_ttd15m\_at\_off+10m}
\item \texttt{d2-d1dep-cartfailure-on-05-easy\_ttd60m\_at\_off+10m}
\item \texttt{d2-cc1-admanualgc-on-03-easy\_ttd480m\_range30m\_off+60m}
\item \texttt{d5-ic2-adfailure-on-04-easy\_ttd480m\_range30m\_off+60m}
\item \texttt{d1-i2-adfailure-on-univ00-universal\_ttd480m\_period\_same-period}
\item \texttt{d5-ic2-imageslowload-on-univ00-universal\_ttd480m\_period\_previous-period}
\item \texttt{d2-d1dep-paymentfailure-50-univ00-universal\_ttd480m\_period\_same-period}
\item \texttt{d3-d2dep-productcatalogfailure-on-univ00-universal\_ttd480m\_at\_off+10m}
\item \texttt{d2-cc1-adfailure-on-univ00-universal\_ttd480m\_at\_off+10m}
\item \texttt{d3-d2dep-paymentunreachable-on-univ00-universal\_ttd15m\_at\_off+10m}
\item \texttt{d5-d4dep-paymentunreachable-on-univ00-universal\_ttd15m\_at\_off+10m}
\item \texttt{d5-d4dep-paymentfailure-10-univ00-universal\_ttd480m\_range30m\_off+10m}
\item \texttt{d1-n1-loadgeneratorfloodhomepage-on-univ00-universal\_ttd60m\_range30m\_off+10m}
\item \texttt{d6-big-cartfailure-on-univ00-universal\_ttd480m\_range30m\_off+60m}
\item \texttt{d4-d3dep-paymentfailure-50-univ00-universal\_ttd60m\_range30m\_off+10m}
\item \texttt{d4-d3dep-paymentfailure-10-01-hard\_ttd1440m\_day\_same-day}
\item \texttt{d2-cc1-adfailure-on-01-hard\_ttd1440m\_day\_same-day}
\item \texttt{d2-d1dep-cartfailure-on-00-hard\_ttd480m\_period\_same-period}
\item \texttt{d2-d1dep-paymentfailure-50-01-hard\_ttd1440m\_day\_same-day}
\item \texttt{d6-big-paymentfailure-50-01-hard\_ttd480m\_period\_same-period}
\item \texttt{d6-big-productcatalogfailure-on-01-hard\_ttd15m\_at\_off+10m}
\item \texttt{d1-c1-admanualgc-on-00-hard\_ttd60m\_at\_off+10m}
\item \texttt{23c5cb27fee69d1e}
\item \texttt{d4-d3dep-paymentfailure-50-02-hard\_ttd60m\_range30m\_off+10m}
\item \texttt{d3-d2dep-productcatalogfailure-on-00-hard\_ttd480m\_range30m\_off+10m}
\item \texttt{d5-d4dep-paymentunreachable-on-00-hard\_ttd480m\_range30m\_off+60m}
\end{itemize}}

\clearpage
\section{Prompts}
\label{appendix:prompts}

This section reproduces every prompt \system{} sends, in the configuration used in our
experiments. Each prompt is a template; values filled in at run time are shown as
\promptvar{name}, and all other text is verbatim. Every step returns its result by calling a
submit tool whose arguments follow a fixed schema (for example, a hypothesis has a
\texttt{statement}, a \texttt{rationale} and a \texttt{region}; an observation has a
\texttt{region}, the retrieval ids it rests on, and a verdict per hypothesis). Tool
calls interleave with each step until that tool is called.

\subsection{Plan}
\label{appendix:prompt:plan}
One call before the first round. It separates the problem from the specification of the answer, so only the final step is handed formats and output paths.

\promptrole{System}
\begin{lstlisting}[style=prompt]
You read a task statement and separate it into parts. You are not solving it. Copy text; do not paraphrase.

## Parts

- `situation`: the problem, the system, and what can be used to investigate it.
- `deliverable`: the required form of the final answer (sections, formats, where to write it). Empty if none is given.
- `requirements`: what a finished answer must contain, one short concrete line each, taken from the deliverable.
- Every part of the statement goes in exactly one of `situation` and `deliverable`.
\end{lstlisting}

\promptrole{User}
\begin{lstlisting}[style=prompt]
---
(*@\promptvar{task statement}@*)
---
\end{lstlisting}

\subsection{Round conversation (system)}
\label{appendix:prompt:session}
Each round is one conversation. This is its system turn; the \Explore, selection, \GlobalTest and \revise prompts below are its user turns, and the model finishes each by calling that stage's submit tool. Tool calls happen inside the turns.

\promptrole{System}
\begin{lstlisting}[style=prompt]
You are investigating a task in rounds. This conversation is one round: each message gives you the next stage, and you finish a stage by calling its submit tool.

## How a round goes

- Explore: map where answers and evidence could lie, and propose hypotheses.
- Hypotheses are then built and tested one by one, outside this conversation; their results come back to you.
- Global test: gather evidence that tells the tested hypotheses apart.
- Revise: decide, for each tested hypothesis, whether to keep, repair, refine or shelve it.
- The state carried over from earlier rounds is in the first message.
\end{lstlisting}

\promptrole{User}
\begin{lstlisting}[style=prompt]
Round (*@\promptvar{round number}@*).
\end{lstlisting}

\subsection{Explore}
\label{appendix:prompt:explore}
Opens each round's conversation, so it carries the whole state; its two parts are sent together as the conversation's first user turn. Shown for a later round; in round one the map, hypothesis, evidence and shelf sections are absent and the model opens the map from the task alone.

\promptrole{System}
\begin{lstlisting}[style=prompt]
You open lines of investigation. You map where answers and evidence could lie, and propose hypotheses to test. Others build and test them; you do not judge them.

## The map

- Hypothesis regions: kinds of answer, grouped under axes (the dimensions answers differ along).
- Observation regions: kinds of evidence, grouped under axes (where and how one would look).
- Add a region only if it is distinct from every region already on the map. Add an axis only if it is new, or subsumes an existing axis (then set `replaces_axis`).
- You can use the tools to look at the environment before you propose; a hypothesis grounded in what exists beats one about what might.

## Hypotheses

- Propose (*@\promptvar{n}@*) hypotheses in the least explored regions and axes (counts are shown; UNTRIED regions and thin axes first). Each needs a testable `statement`, a `rationale`, and a `region`.
- Make them compete: different regions, different mechanisms, not variants of one idea.
- Read the evidence on last round's hypotheses and every hypothesis's metrics per part of the task. Do not re-propose what is held.
- If hypotheses win on different parts, or the best score has stopped improving, at least one proposal must be a combination (name the hypotheses and what it takes from each) or a `revive` of an eligible shelved hypothesis.
- Call submit_exploration once.
\end{lstlisting}

\promptrole{User}
\begin{lstlisting}[style=prompt]
**Task**

(*@\promptvar{task}@*)

(*@\promptvar{starting artifact, if any}@*)

**Hypothesis regions**

(*@\promptvar{hypothesis regions, by axis, with counts}@*)

**Observation regions**

(*@\promptvar{observation regions, by axis, with counts}@*)

**Current hypotheses**

(*@\promptvar{hypotheses with status, scores and evidence counts}@*)

**Evidence on last round's hypotheses**

(*@\promptvar{evidence on last round's hypotheses}@*)

**Observations**

(*@\promptvar{observations}@*)

**Shelved, that you should consider for reuse especially if the scores plateauing. It could be useful to take different ideas from different hypothesis and create a hypothesis that combines them.**

(*@\promptvar{shelved hypotheses past their cooldown}@*)

**Shelved, but different from the leader on some part of the task** (not revivable yet; their artifacts can be ingredients of a new hypothesis that combines them)

(*@\promptvar{shelved hypotheses that win on some part of the task}@*)
\end{lstlisting}

\subsection{Select}
\label{appendix:prompt:session_select}
From round two on, chooses which new, refined, repaired or revived hypotheses are tested. Later turns send only what changed since the model last saw it.

\promptrole{User}
\begin{lstlisting}[style=prompt]
Next stage: select.

## Choose

- Candidates: (*@\promptvar{pending hypothesis ids}@*). Choose at most (*@\promptvar{k}@*) to build and test this round; the rest are shelved and can be revived later.
- Kept hypotheses ((*@\promptvar{kept hypothesis ids}@*)) carry over untested and take no slot.
- Prefer tests that would settle what the evidence leaves open; spread across regions unless the evidence clearly favours one.
- Call submit_selection once.

## New since your last message

**Hypotheses**

(*@\promptvar{new or changed hypotheses}@*)

**Observations**

(*@\promptvar{new observations}@*)
\end{lstlisting}

\subsection{Candidate analysis (local test)}
\label{appendix:prompt:local}
One per selected hypothesis, run by a separate tool-using agent outside the round conversation; this is one iteration of its build--evaluate--learn loop. Shown with every optional section; the artifact and evaluator lines appear only on benchmarks that execute candidates.

\promptrole{User}
\begin{lstlisting}[style=prompt]
You test one hypothesis. This is iteration (*@\promptvar{i}@*) of at most (*@\promptvar{max iterations}@*); each iteration builds on the last.

## This iteration

- Build an artifact that expresses the hypothesis, write it to `(*@\promptvar{artifact path}@*)`, and run it. (*@\promptvar{artifact contract}@*)
- Measure it with `(*@\promptvar{evaluator tool}@*)`, once, on the file you wrote, and wait for the result: evaluations take turns, so a call can wait many minutes before it starts. Improve on the previous iteration where you can; a lower score is still a result.
- Look for evidence that supports and that refutes the hypothesis. The observation regions below are kinds of evidence to look for, not tools.
- Revise the claim only if evidence you gathered contradicts it, or a small change of approach -- not a code fix -- should help, and the revision predicts a better score (`predicts_improvement`). A finding that something does not help is an observation refuting the claim, not a revision.

## What to submit

- `observations`: what you found, each with its region, the retrieval ids it rests on, and a verdict on (*@\promptvar{hypothesis id}@*).
- Separate a flaw in the artifact from a flaw in the claim; a crash is not a refutation.
- Set `done` when another iteration would not add anything. End with submit_local: your final answer in its shape.

## Context

**Task**

(*@\promptvar{task}@*)

**Hypothesis under test**

(*@\promptvar{hypothesis, with its record}@*)

(*@\promptvar{starting artifact}@*) Gains are measured against it.

**Start from this artifact**: `(*@\promptvar{base artifact path}@*)` ((*@\promptvar{its origin}@*)). Copy it to your artifact path and edit the copy; where it marks an editable region (for example between EVOLVE-BLOCK markers), change only what is inside that region.

**Repair requested** (keep the claim, fix how it is realized): (*@\promptvar{what the revision asked to fix}@*)

**Observation regions** (kinds of evidence)

(*@\promptvar{observation regions}@*)

**Previous iterations**

- iteration (*@\promptvar{i-1}@*): score (*@\promptvar{score}@*); (*@\promptvar{artifact path}@*); (*@\promptvar{notes}@*)
\end{lstlisting}

\subsection{Comparative analysis (global test)}
\label{appendix:prompt:session_global_test}
After the local tests return, looks for evidence that tells the tested hypotheses apart.

\promptrole{User}
\begin{lstlisting}[style=prompt]
Next stage: global test. (*@\promptvar{tested hypothesis ids}@*) are tested; the local results that came back are below.

## Read the local results with care

- Scores and support/refute counts come from experiments that can themselves be flawed: a bug, a bad measurement, a narrow check.
- Look at the artifacts and the evidence behind each result before trusting it.
- Where two hypotheses predict different things, find out which happened. Compare their metrics part by part: a claim that loses overall can win somewhere.

## What to submit

- New observations, each with its observation region and the retrieval ids it rests on.
- For each, a verdict (supports, refutes, neutral) on every hypothesis it bears on, with one line why.
- Call submit_evidence once.

## New since your last message

**Hypotheses**

(*@\promptvar{hypotheses changed by the local tests}@*)

**Observations**

(*@\promptvar{observations filed by the local tests}@*)
\end{lstlisting}

\subsection{Exploit (revise)}
\label{appendix:prompt:session_revise}
One decision per tested hypothesis: keep, repair, refine or shelve (set aside). \texttt{repair} is offered only on benchmarks that execute candidates.

\promptrole{User}
\begin{lstlisting}[style=prompt]
Next stage: revise. Decide every one of: (*@\promptvar{tested hypothesis ids}@*).

## Decisions

- `keep`: the claim stands; it is carried forward without retesting.
- `repair`: the claim is sound but its artifact expressed it badly; it is rebuilt and retested.
- `refine`: the claim needs a changed clause, scope or mechanism; give the new `statement`.
- `shelve`: set it aside; it can be revived if the run stops making progress.

## How to decide

- A result that rests on one measurement of one artifact is weak evidence either way.
- Do not shelve a claim that wins on some part of the task, even if it loses overall: say so in `rationale`.
- Weigh what supports each claim against what refutes it, and cite those observation ids in `rationale`; your last submission was filed as the observations below.
- Call submit_revision once.

## New since your last message

(*@\promptvar{observations from the global test}@*)

(*@\promptvar{changed hypotheses}@*)

**Evidence on each hypothesis to decide**

(*@\promptvar{supporting and refuting observations per hypothesis}@*)
\end{lstlisting}

\subsection{Final answer}
\label{appendix:prompt:finalize}
After the last round, whatever the budget.

\promptrole{System}
\begin{lstlisting}[style=prompt]
You write the final answer from an investigation that is now over.

## How

- Build the answer from the best-supported hypotheses and from what the observations ruled out.
- Cite the evidence the answer rests on where the format asks for it; the retrieval ids and their contents are below.
- Do not state as established what was only not refuted.
- If the answer is an artifact, give the path of the best one in `file_path`. (*@\promptvar{artifact contract}@*)
- Call submit_final once.

## Format

(*@\promptvar{the benchmark's answer format}@*)
\end{lstlisting}

\promptrole{User}
\begin{lstlisting}[style=prompt]
**Task**

(*@\promptvar{task statement}@*)

**The answer must contain**

- (*@\promptvar{requirement}@*)

**Hypotheses, best first**

(*@\promptvar{hypotheses, ranked, with scores and evidence}@*)

**Observations**

(*@\promptvar{observations}@*)

**Retrievals**

(*@\promptvar{retrieval index and contents}@*)
\end{lstlisting}

\section{State Representation}
\label{appendix:state}

\system{} keeps its state $S_t$ in three record types: regions, hypotheses and observations. Nothing in the
state calls a model; stages read it through fixed renderings and change it only
through the operations below, so every change is recorded in the trace.

\begin{lstlisting}[language=Python, basicstyle=\ttfamily\scriptsize, columns=fullflexible, keepspaces=true, frame=single, rulecolor=\color{black!30}, backgroundcolor=\color{black!3}, commentstyle=\color{black!55}\itshape, keywordstyle=\bfseries, morekeywords={dataclass}]
@dataclass
class Region:
    axis: str            # the dimension this region splits along
    name: str            # identity of the region within its space
    description: str

@dataclass
class Hypothesis:
    id: str              # h1, h2, ...
    statement: str       # the testable claim
    rationale: str
    region: str          # a hypothesis region
    status: str          # pending | tested | shelved
    parent: str          # set on a refinement: the hypothesis it refines
    born: int            # round proposed
    touched: int         # last round it was proposed, tested, shelved or revived
    reason: str          # why it was shelved
    repair: str          # what a requested repair should fix
    artifact: str        # best artifact its local test produced
    scores: list[float]  # every evaluator score of its artifacts
    metrics: dict        # per-part metrics of its best artifact
    notes: str           # the local test's own summary

@dataclass
class Observation:
    id: str              # o1, o2, ...
    content: str         # what was found, checkable by another reader
    region: str          # an observation region
    sources: list[str]   # ids of the tool results it rests on
    round: int           # round filed
    origin: str          # the step that filed it
    verdicts: dict       # hypothesis id -> {relation, why}; supports | refutes | neutral
    count: str           # how many experiments have run into the same conclusion
\end{lstlisting}

\paragraph{Axes and regions.} The hypothesis space $H$ and the observation space $O$
are each a map from region name to \texttt{Region}; each region belongs to one axis,
so a space is a two-level partition: axes (the dimensions along which answers, or
ways of looking, differ) and regions within them (kinds of answer, or kinds of
evidence). \Explore declares both. A region whose name matches
an existing one after normalisation is folded into it. A new axis
that subsumes an existing one is declared with \texttt{replaces\_axis}, and the old
axis's regions move under it. Every hypothesis names one hypothesis region and every
observation one observation region.

\paragraph{Hypotheses.} The working hypothesis set $wH$ is the hypotheses whose status
is \texttt{pending} (awaiting \LocalTest: new, refined, repaired or revived) or
\texttt{tested} (tested and \retained, carried forward without retesting); a
\texttt{shelved} hypothesis is \setaside. Nothing is deleted. A claim identical to
one already held after normalisation is not added again. A refinement is a new hypothesis in the same region with \texttt{parent} set, so the trace
shows every claim in the form it was tested.

\paragraph{Observations and evidence.} Observations are filed by \LocalTest and
\GlobalTest. An observation whose content repeats an existing one is folded into it:
its sources are merged and its verdicts added. The evidence relation $E$ is conceptually separate but, at the data level, is stored on
the observations as \texttt{verdicts}: for each hypothesis the observation bears on,
a relation (\textsc{Support}, \textsc{Refute} or \textsc{Neutral}) and a one-line
reason; a pair with no verdict is unresolved. The support and refute counts shown
for a hypothesis are counts over these verdicts.

\paragraph{What the stages see.} Each space is rendered axis by axis, with counts per
axis and per region: for hypothesis regions, how many hypotheses were proposed and
tested there and the best score reached; for observation regions, how many
observations were filed. A hypothesis region with nothing proposed in it is marked
\textsc{Untried}. These counts are what \Explore is asked to use when it proposes in
the least explored regions and axes. A hypothesis is rendered with its status,
region, parent, scores, evidence counts, per-part metrics and artifact; an
observation with its region, sources and verdicts; and \revise additionally sees,
for each hypothesis it decides, the observations that support and refute it.

\section{Coverage Analysis}
\label{appendix:coverage}

This section gives the procedure behind the breadth measurements of Section~\ref{sec:breadth}, which follow the coverage analysis of AIM~\citep{choi2026aim} with two extensions: a sample-size-matched hull and deduplication.

\paragraph{What counts as a solution.} A solution is any complete candidate a method submitted to the task's evaluator during a run, whether or not it became the final answer. For the optimization benchmarks (QNN, LLM\_SQL, PRISM, Cloudcast, EPLB, transaction scheduling) it is the program text passed to the evaluator; for ARC-AGI-2 it is the transform function of every sandbox cell that defined one, extracted from the cell's source by parsing rather than pattern matching, so helper code and the test harness around it are not embedded; for DOCKSTRING it is the SMILES string of every molecule docked. For EvoX the solutions are the programs in the final checkpoint's database (every program it evaluated), ordered by the iteration that found them, or its docking log. The single-agent baselines contribute the candidates they evaluated along the way; a baseline that evaluates nothing before answering has one solution and no hull.

\paragraph{Distinct solutions.} Identical texts within a run are counted once before anything is embedded. A loop that re-evaluates the same program, or a baseline that docks the same molecule under two names, is not exploring, and the hull would otherwise count the repeat as a point. On PRISM the direct baseline's 94 evaluations were 18 distinct programs; on DOCKSTRING EvoX's 494 dockings were 80 distinct molecules.

\paragraph{Embedding.} Programs are embedded with \texttt{text-embedding-3-large} (3,072 dimensions). The model accepts 8,191 tokens, so each text is cut at 24,000 characters. The cut touches none of \system{}'s or the single-agent baselines' programs; it touches EvoX's, which carry the whole task file (142 of 166 on LLM\_SQL, 79 of 186 on Cloudcast, 38 of 157 on transaction scheduling, 25 of 182 on PRISM, none on QNN or EPLB). No two of those programs agree on their first 24,000 characters, so the cut drops tails without merging programs, but EvoX's spread on those tasks is measured on the first 24,000 characters of each program. Molecules are not embedded with a text model, since an English-text embedding of a SMILES string measures its spelling; we use 2,048-bit Morgan fingerprints of radius 2, the standard representation of molecular neighbourhoods, and all downstream geometry is computed on those bit vectors.

\paragraph{Projection.} For each task we pool the distinct solutions of every method and run, center them, and take the first two right singular vectors of the pooled matrix as the task's plane. One plane per task, shared by all methods, is what makes hull areas comparable within a task: projecting each run onto its own principal components would give every run the axes along which it happens to vary most. Areas are in the units of that plane and are not comparable across tasks, which is why every figure is per task.

\paragraph{Hull area.} The convex hull of a run's projected points is computed with the monotone-chain algorithm and its area with the shoelace formula; fewer than three distinct points give zero.

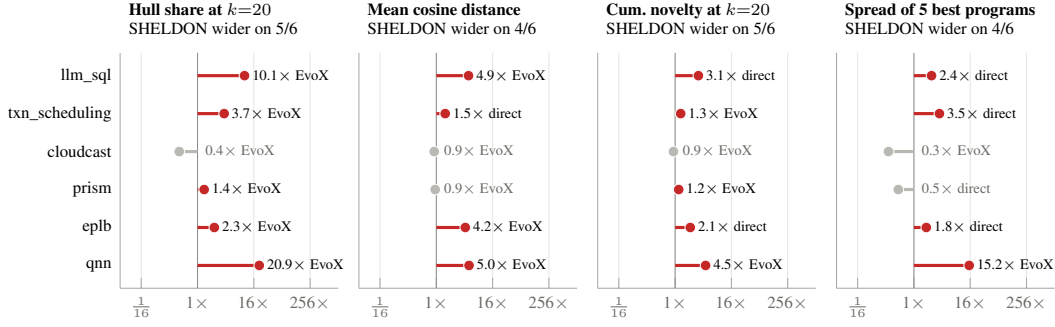
\begin{figure}[tp]
    \centering
\resizebox{\linewidth}{!}{%
\begin{tikzpicture}
\begin{groupplot}[sheldonfig,
  group style={group size=4 by 1, horizontal sep=0.3cm, yticklabels at=edge left},
  width=4.75cm, height=5.0cm, xmin=-5.5, xmax=10, clip=false, ytick style={draw=none}, ymin=-0.6, ymax=5.6,
  ytick={0,...,5}, yticklabels={qnn,eplb,prism,cloudcast,txn\_scheduling,llm\_sql},
  yticklabel style={font=\scriptsize, text=black},
  xtick={-4,0,4,8}, xticklabels={$\frac{1}{16}$,1$\times$,16$\times$,256$\times$},
  ymajorgrids=false, title style={font=\scriptsize\bfseries}]
\nextgroupplot[title={Hull share at $k{=}20$\\\textmd{SHELDON wider on 5/6}}]
\draw[black!45] (axis cs:0,-0.6) -- (axis cs:0,5.6);
\draw[sheldon, line width=1.3pt] (axis cs:0,5) -- (axis cs:3.339,5);
\fill[sheldon, draw=white, line width=0.4pt] (axis cs:3.339,5) circle (2.2pt);
\node[anchor=west, font=\tiny, inner sep=1pt, text=black] at (axis cs:3.689,5) {10.1$\times$ EvoX};
\draw[sheldon, line width=1.3pt] (axis cs:0,4) -- (axis cs:1.893,4);
\fill[sheldon, draw=white, line width=0.4pt] (axis cs:1.893,4) circle (2.2pt);
\node[anchor=west, font=\tiny, inner sep=1pt, text=black] at (axis cs:2.243,4) {3.7$\times$ EvoX};
\draw[neutral, line width=1.3pt] (axis cs:0,3) -- (axis cs:-1.299,3);
\fill[neutral, draw=white, line width=0.4pt] (axis cs:-1.299,3) circle (2.2pt);
\node[anchor=west, font=\tiny, inner sep=1pt, text=muted] at (axis cs:0.350,3) {0.4$\times$ EvoX};
\draw[sheldon, line width=1.3pt] (axis cs:0,2) -- (axis cs:0.472,2);
\fill[sheldon, draw=white, line width=0.4pt] (axis cs:0.472,2) circle (2.2pt);
\node[anchor=west, font=\tiny, inner sep=1pt, text=black] at (axis cs:0.822,2) {1.4$\times$ EvoX};
\draw[sheldon, line width=1.3pt] (axis cs:0,1) -- (axis cs:1.194,1);
\fill[sheldon, draw=white, line width=0.4pt] (axis cs:1.194,1) circle (2.2pt);
\node[anchor=west, font=\tiny, inner sep=1pt, text=black] at (axis cs:1.544,1) {2.3$\times$ EvoX};
\draw[sheldon, line width=1.3pt] (axis cs:0,0) -- (axis cs:4.388,0);
\fill[sheldon, draw=white, line width=0.4pt] (axis cs:4.388,0) circle (2.2pt);
\node[anchor=west, font=\tiny, inner sep=1pt, text=black] at (axis cs:4.738,0) {20.9$\times$ EvoX};
\nextgroupplot[title={Mean cosine distance\\\textmd{SHELDON wider on 4/6}}]
\draw[black!45] (axis cs:0,-0.6) -- (axis cs:0,5.6);
\draw[sheldon, line width=1.3pt] (axis cs:0,5) -- (axis cs:2.289,5);
\fill[sheldon, draw=white, line width=0.4pt] (axis cs:2.289,5) circle (2.2pt);
\node[anchor=west, font=\tiny, inner sep=1pt, text=black] at (axis cs:2.639,5) {4.9$\times$ EvoX};
\draw[sheldon, line width=1.3pt] (axis cs:0,4) -- (axis cs:0.631,4);
\fill[sheldon, draw=white, line width=0.4pt] (axis cs:0.631,4) circle (2.2pt);
\node[anchor=west, font=\tiny, inner sep=1pt, text=black] at (axis cs:0.981,4) {1.5$\times$ direct};
\draw[neutral, line width=1.3pt] (axis cs:0,3) -- (axis cs:-0.151,3);
\fill[neutral, draw=white, line width=0.4pt] (axis cs:-0.151,3) circle (2.2pt);
\node[anchor=west, font=\tiny, inner sep=1pt, text=muted] at (axis cs:0.350,3) {0.9$\times$ EvoX};
\draw[neutral, line width=1.3pt] (axis cs:0,2) -- (axis cs:-0.081,2);
\fill[neutral, draw=white, line width=0.4pt] (axis cs:-0.081,2) circle (2.2pt);
\node[anchor=west, font=\tiny, inner sep=1pt, text=muted] at (axis cs:0.350,2) {0.9$\times$ EvoX};
\draw[sheldon, line width=1.3pt] (axis cs:0,1) -- (axis cs:2.060,1);
\fill[sheldon, draw=white, line width=0.4pt] (axis cs:2.060,1) circle (2.2pt);
\node[anchor=west, font=\tiny, inner sep=1pt, text=black] at (axis cs:2.410,1) {4.2$\times$ EvoX};
\draw[sheldon, line width=1.3pt] (axis cs:0,0) -- (axis cs:2.325,0);
\fill[sheldon, draw=white, line width=0.4pt] (axis cs:2.325,0) circle (2.2pt);
\node[anchor=west, font=\tiny, inner sep=1pt, text=black] at (axis cs:2.675,0) {5.0$\times$ EvoX};
\nextgroupplot[title={Cum.\ novelty at $k{=}20$\\\textmd{SHELDON wider on 5/6}}]
\draw[black!45] (axis cs:0,-0.6) -- (axis cs:0,5.6);
\draw[sheldon, line width=1.3pt] (axis cs:0,5) -- (axis cs:1.655,5);
\fill[sheldon, draw=white, line width=0.4pt] (axis cs:1.655,5) circle (2.2pt);
\node[anchor=west, font=\tiny, inner sep=1pt, text=black] at (axis cs:2.005,5) {3.1$\times$ direct};
\draw[sheldon, line width=1.3pt] (axis cs:0,4) -- (axis cs:0.396,4);
\fill[sheldon, draw=white, line width=0.4pt] (axis cs:0.396,4) circle (2.2pt);
\node[anchor=west, font=\tiny, inner sep=1pt, text=black] at (axis cs:0.746,4) {1.3$\times$ EvoX};
\draw[neutral, line width=1.3pt] (axis cs:0,3) -- (axis cs:-0.118,3);
\fill[neutral, draw=white, line width=0.4pt] (axis cs:-0.118,3) circle (2.2pt);
\node[anchor=west, font=\tiny, inner sep=1pt, text=muted] at (axis cs:0.350,3) {0.9$\times$ EvoX};
\draw[sheldon, line width=1.3pt] (axis cs:0,2) -- (axis cs:0.256,2);
\fill[sheldon, draw=white, line width=0.4pt] (axis cs:0.256,2) circle (2.2pt);
\node[anchor=west, font=\tiny, inner sep=1pt, text=black] at (axis cs:0.606,2) {1.2$\times$ EvoX};
\draw[sheldon, line width=1.3pt] (axis cs:0,1) -- (axis cs:1.077,1);
\fill[sheldon, draw=white, line width=0.4pt] (axis cs:1.077,1) circle (2.2pt);
\node[anchor=west, font=\tiny, inner sep=1pt, text=black] at (axis cs:1.427,1) {2.1$\times$ direct};
\draw[sheldon, line width=1.3pt] (axis cs:0,0) -- (axis cs:2.165,0);
\fill[sheldon, draw=white, line width=0.4pt] (axis cs:2.165,0) circle (2.2pt);
\node[anchor=west, font=\tiny, inner sep=1pt, text=black] at (axis cs:2.515,0) {4.5$\times$ EvoX};
\nextgroupplot[title={Spread of 5 best programs\\\textmd{SHELDON wider on 4/6}}]
\draw[black!45] (axis cs:0,-0.6) -- (axis cs:0,5.6);
\draw[sheldon, line width=1.3pt] (axis cs:0,5) -- (axis cs:1.254,5);
\fill[sheldon, draw=white, line width=0.4pt] (axis cs:1.254,5) circle (2.2pt);
\node[anchor=west, font=\tiny, inner sep=1pt, text=black] at (axis cs:1.604,5) {2.4$\times$ direct};
\draw[sheldon, line width=1.3pt] (axis cs:0,4) -- (axis cs:1.814,4);
\fill[sheldon, draw=white, line width=0.4pt] (axis cs:1.814,4) circle (2.2pt);
\node[anchor=west, font=\tiny, inner sep=1pt, text=black] at (axis cs:2.164,4) {3.5$\times$ direct};
\draw[neutral, line width=1.3pt] (axis cs:0,3) -- (axis cs:-1.807,3);
\fill[neutral, draw=white, line width=0.4pt] (axis cs:-1.807,3) circle (2.2pt);
\node[anchor=west, font=\tiny, inner sep=1pt, text=muted] at (axis cs:0.350,3) {0.3$\times$ EvoX};
\draw[neutral, line width=1.3pt] (axis cs:0,2) -- (axis cs:-1.109,2);
\fill[neutral, draw=white, line width=0.4pt] (axis cs:-1.109,2) circle (2.2pt);
\node[anchor=west, font=\tiny, inner sep=1pt, text=muted] at (axis cs:0.350,2) {0.5$\times$ direct};
\draw[sheldon, line width=1.3pt] (axis cs:0,1) -- (axis cs:0.881,1);
\fill[sheldon, draw=white, line width=0.4pt] (axis cs:0.881,1) circle (2.2pt);
\node[anchor=west, font=\tiny, inner sep=1pt, text=black] at (axis cs:1.231,1) {1.8$\times$ direct};
\draw[sheldon, line width=1.3pt] (axis cs:0,0) -- (axis cs:3.929,0);
\fill[sheldon, draw=white, line width=0.4pt] (axis cs:3.929,0) circle (2.2pt);
\node[anchor=west, font=\tiny, inner sep=1pt, text=black] at (axis cs:4.279,0) {15.2$\times$ EvoX};
\end{groupplot}
\end{tikzpicture}%
}
    \caption{Ratio of \system{}'s median to the strongest baseline's median on each breadth metric, log scale; the baseline is named beside each point. Blue: \system{} is wider; grey: a baseline is.}
\label{fig:explore-summary}
\end{figure}

\paragraph{Hull area at matched sample size.} A convex hull can only grow as points are added, so the raw hull credits a method for having evaluated more. For a sample size $k$ we draw $k$ of each run's distinct solutions uniformly without replacement, compute the hull of the draw, and average over 300 draws with a fixed seed. Figure~\ref{fig:matched-coverage} reports this for every $k$ up to each run's size, as a share of the hull of all distinct solutions on the task, and the summary in Figure~\ref{fig:explore-summary} reads it at $k{=}20$.

\paragraph{Cosine diversity.} For each run we report the mean pairwise cosine distance $D = 1-\cos\theta$ over all pairs of its distinct solutions' embeddings, in the full embedding space rather than the plane. It is independent of sample size in expectation and serves as a check on the hull.

\paragraph{Cumulative novelty.} For a run's distinct solutions in evaluation order, the novelty of each solution is its distance in the full embedding space to the nearest solution evaluated before it; cumulative novelty after $k$ solutions is the sum of the first $k$ novelties. A method that keeps proposing variants of what it already has accumulates novelty slowly.

\paragraph{Distinct strong solutions.} For each run with at least five valid programs we report the mean pairwise distance, in the task's plane, among its five highest-scoring valid programs, and the number of cells of an $8{\times}8$ grid over the plane that contain at least one program scoring in the top quartile of all valid programs evaluated on the task, pooled across methods.

\paragraph{Distinct ideas on text tasks.} On DiagnosisArena and ORCA-Bench a hypothesis is text rather than a program, so breadth is measured on ideas instead of embeddings of programs. Each hypothesis on which \system{} ran a candidate analysis, and each frontier to which GoS dispatched an expert, is reduced to a short diagnosis or root-cause label by \texttt{gpt-5-mini} under a fixed instruction, and the label is embedded. Labels are read in exploration order, and a label counts as a new idea if its cosine distance to every earlier label of that run exceeds a radius $r$; we report $r \in \{0.25, 0.35, 0.5\}$. Tasks for which a system produced no labels are omitted.

\paragraph{What the plane shows and does not.} The two components retain a fraction of the variance that varies by task, and a hull in the plane is a proxy for breadth rather than a measurement of it; $D$ is reported beside it for that reason. Both numbers describe where a method looked, not what it found, so they are read together with the scores of Section~\ref{sec:experiments}, and neither accounts for cost.

\paragraph{Structural quantities.} The quantities of Section~\ref{sec:obs-contribution} are computed from each run's recorded hypothesis and observation spaces. The winner of a run is the hypothesis whose artifact the final step submitted; when it did not name one, the best-measured hypothesis; otherwise the one with the most net support. Observations filed under a region name that was never registered, or under an axis name rather than a region, are kept under that name as their own row rather than discarded, so counts are over everything the run wrote. For the earlier two-space loop used on QNN and the ADRS tasks, the observation space's categories stand in for observation regions and there are no hypothesis regions; per-round scores come from the evaluator's own record rather than from the hypotheses.

\textbf{Across metrics, \system{} is wider in 18 of 24 task--metric pairs.} Figure~\ref{fig:explore-summary} summarizes four breadth metrics as the ratio of \system{}'s median to the strongest baseline's median: hull share at $k{=}20$; cosine diversity, the mean pairwise cosine distance $D = 1-\cos\theta$ over all pairs of a run's distinct solutions, computed in the full embedding space rather than the plane; cumulative novelty after 20 programs; and spread of the five best programs. Cosine diversity is AIM's supplementary number and is independent of sample size in expectation, so it checks the hull: a run can have a large hull from three outliers and a small $D$, or a small hull and a large $D$ when its variation lies outside the two retained components. Both describe where a method looked, not what it found, and neither accounts for cost; the cost-matched comparisons are in Section~\ref{sec:experiments}. Cloudcast is the consistent exception: its best solutions sit in a small part of the space, and \system{} converges there.

\section{Cost}
\label{appendix:cost}

Table~\ref{tab:cost} reports what each method spent, and how long it took, per task on DiagnosisArena and ORCA-Bench.
All costs are computed from each run's token counts under one pricing rule, the list
price of GPT-5.6 Terra (\$2.50 per million input tokens, \$0.25 per million cached input
tokens, \$15.00 per million output tokens).

All methods are run with a budget limit of \$10 per task. \system{} spends more than the single-agent baselines because it is more meticulous.
Diagnosing a patient and finding the root cause of an incident are tasks where a
confident wrong answer is costly: rather than committing to the first
plausible explanation, it keeps several hypotheses open, tests each one against the
evidence, and compares them before answering. Each round therefore runs several
investigations.
The comparison with GoS, which also maintains multiple hypotheses and spends a similar
amount, shows that the cost alone does not buy the accuracy: at a comparable budget,
\system{} scores higher on both benchmarks.

\begin{table}[ht]
\centering\small
\caption{Cost (\$) and wall-clock time (minutes) per task, on the 50 DiagnosisArena cases
and the 30 ORCA-Bench incidents.}
\label{tab:cost}
\begin{tabular}{@{}l r r r r@{}}
\toprule
& \multicolumn{2}{c}{DiagnosisArena} & \multicolumn{2}{c}{ORCA-Bench} \\
\cmidrule(lr){2-3}\cmidrule(l){4-5}
Method & Cost & Time & Cost & Time \\
\midrule
ReAct     & 0.02 & 0.3  & 0.54 & 4.2 \\
Codex     & 0.06 & 0.8  & 0.73 & 4.5 \\
RLM       & 0.09 & 1.7  & 1.56 & 2.7 \\
GoS       & 2.79 & 29.1 & 7.54 & 55.4 \\
\system{} & 2.91 & 14.0 & 6.90 & 22.0 \\
\bottomrule
\end{tabular}
\end{table}

\section{Experiment Configuration}
\label{appendix:configs}

A run is specified by one configuration file. We show the configuration of one ADRS
task (Cloudcast) as a representative; runs on other benchmarks
differ in the benchmark options, the budget, the round cap, and otherwise use the same method hyperparameters. \texttt{max\_usd} is the
budget per task in US dollars; \texttt{null} lifts a cap.
Table~\ref{tab:config-method} explains the method hyperparameters.

\begin{lstlisting}[basicstyle=\ttfamily\scriptsize, columns=fullflexible, keepspaces=true, breaklines=true, frame=single, rulecolor=\color{black!30}, backgroundcolor=\color{black!3}]
benchmark: skydiscover
split: adrs
method: hypodeducto
model: gpt-5.6-terra
effort: high
max_usd: 15
max_rounds: null
max_calls: null
params:
  mode: session
  plan: true
  rounds: 8
  n_initial: 5
  n_explore: 3
  select_k: 5
  local_max_evals: 3
  local_patience: 2
  local_min_gain_pct: 0.5
  shelf_cooldown: 3
  stagnation_rounds: 2
  round_reserve: 0.5
  explore_probe_calls: 20
  select_probe_calls: 2
  global_test_probe_calls: 10
  revise_probe_calls: 4
  finalize_probe_calls: 20
  local_concurrent: true
  local_probe_calls: 20
  local_engine: toolloop
  local_engine_params: {model: gpt-5.6-luna}
  probes: [run_shell]
  benchmark_opts: {tasks: ADRS.cloudcast, isolate: true, runtime: host, shell_nice: 10, state_timeout: true}
\end{lstlisting}

\begin{table}[ht]
\centering\scriptsize
\caption{Method hyperparameters.}
\label{tab:config-method}
\begin{tabular}{@{}l c p{0.58\linewidth}@{}}
\toprule
Parameter & Value & Meaning \\
\midrule
\texttt{mode} & session & a round's deciding steps run as turns of one conversation (\texttt{session}) \\
\texttt{plan} & on & separate the problem from the answer specification before round 1 \\
\texttt{rounds} & 8 & round cap \\
\texttt{n\_initial} & 5 & hypotheses proposed by \Explore in round 1 \\
\texttt{n\_explore} & 3 & hypotheses proposed by \Explore in each later round \\
\texttt{select\_k} & 5 & hypotheses selected for \LocalTest per round (round 2 on) \\
\texttt{local\_max\_evals} & 3 & iterations of one \LocalTest loop (evaluator calls, where there is an evaluator) \\
\texttt{local\_patience} & 2 & scored iterations without improvement that end a \LocalTest loop \\
\texttt{local\_min\_gain\_pct} & 0.5 & improvement, in \% of the starting score, that counts as progress \\
\texttt{shelf\_cooldown} & 3 & rounds before a set-aside hypothesis can be revived \\
\texttt{stagnation\_rounds} & 2 & rounds with no new best and no new observation after which any set-aside hypothesis can be revived \\
\texttt{round\_reserve} & 0.5 & no new round starts with less budget left than this fraction of the mean round cost \\
\texttt{explore\_probe\_calls} & 20 & tool calls allowed in \Explore \\
\texttt{select\_probe\_calls} & 2 & tool calls allowed in selection \\
\texttt{global\_test\_probe\_calls} & 10 & tool calls allowed in \GlobalTest \\
\texttt{revise\_probe\_calls} & 4 & tool calls allowed in \revise \\
\texttt{finalize\_probe\_calls} & 20 & tool calls allowed in the final answer \\
\bottomrule
\end{tabular}
\end{table}

\end{document}